\documentclass[final,5p,times,twocolumn]{elsarticle}

\usepackage{amssymb}
\usepackage{amsmath}
\usepackage{algorithmic}
\usepackage{lineno}
\usepackage{url}
\usepackage{hyperref}
\usepackage{multirow}
\usepackage{booktabs}
\usepackage{makecell}
\usepackage{pifont} 
\usepackage{ulem}
\usepackage{comment}
\usepackage{algorithm}
\usepackage{algorithmic}
\usepackage{xcolor} 
\usepackage{colortbl}
\usepackage{longtable}
\usepackage{amsfonts}  
\usepackage{bbm}  

\journal{xxx}

\begin{document}

\begin{frontmatter}



\title{BleedOrigin: Dynamic Bleeding Source Localization in Endoscopic Submucosal Dissection via Dual-Stage Detection and Tracking\tnoteref{fund}}

\tnotetext[fund]{This work was supported by Hong Kong Research Grants Council (RGC) Collaborative Research Fund (C4026-21G), General Research Fund (GRF 14211420 \& 14203323), Shenzhen-Hong Kong-Macau Technology Research Programme (Type C) STIC Grant SGDX20210823103535014 (202108233000303).}

\author[1,2]{Mengya Xu\fnref{equal}} 
\author[2,4]{Rulin Zhou\fnref{equal}}
\author[1]{An Wang\fnref{equal}}
\author[3]{Chaoyang Lyu}
\author[3]{Zhen Li}
\author[3]{Ning Zhong}
\author[1,2]{Hongliang Ren\corref{cor1}}

\fntext[equal]{Equal Contribution. Mengya Xu is the project leader.}
\cortext[cor1]{Corresponding Author. \ead{hlren@ee.cuhk.edu.hk.}}
\affiliation[1]{organization={Department of Electronic Engineering, The Chinese University of Hong Kong},
            state={Hong Kong SAR},
            country={China}}
\affiliation[2]{organization={The Chinese University of Hong Kong Shenzhen Research Institute},
            city={Shen Zhen},
            state={Guangdong},
            country={China}}
\affiliation[3]{organization={Department of Gastroenterology, Qilu Hospital of Shandong University},
            city={Jinan},
            state={Shandong},
            country={China}} 
            
\affiliation[4]{organization={Department of Mechanical Engineering, The University of Hong Kong},
            state={Hong Kong SAR},
            country={China}}             

\begin{abstract}

Intraoperative bleeding during Endoscopic Submucosal Dissection (ESD) poses significant risks, demanding precise, real-time localization and continuous monitoring of the bleeding source for effective hemostatic intervention. In particular, endoscopists have to repeatedly flush to clear blood, allowing only milliseconds to identify bleeding sources, an inefficient process that prolongs operations and elevates patient risks. However, current Artificial Intelligence (AI) methods primarily focus on bleeding region segmentation, overlooking the critical need for accurate bleeding source detection and temporal tracking in the challenging ESD environment, which is marked by frequent visual obstructions and dynamic scene changes. This gap is widened by the lack of specialized datasets, hindering the development of robust AI-assisted guidance systems. To address these challenges, we introduce \textbf{BleedOrigin-Bench}, the first comprehensive ESD bleeding source dataset, featuring 1,771 expert-annotated bleeding sources across 106,222 frames from 44 procedures, supplemented with 39,755 pseudo-labeled frames. This benchmark covers 8 anatomical sites and 6 challenging clinical scenarios. We also present \textbf{BleedOrigin-Net}, a novel dual-stage detection-tracking framework for the bleeding source localization in ESD procedures, addressing the complete workflow from bleeding onset detection to continuous spatial tracking. For initial detection, our method integrates a Multi-Domain Confidence-based Frame Memory (MDCFM) module that leverages RGB, HSV, and optical flow features for robust temporal context, combined with Multi-Domain Gated Attention (MDG) for superior onset detection. For continuous tracking, we employ a pseudo-label enhanced strategy that incorporates feature matching, trajectory prediction, and Kalman filtering to generate dense supervision from sparse annotations, complemented by parameter-efficient LoRA fine-tuning. We compare with widely-used object detection models (YOLOv11/v12), multimodal large language models, and point tracking methods. Extensive evaluation demonstrates state-of-the-art performance, achieving 96.85\% frame-level accuracy ($\pm\leq8$ frames) for bleeding onset detection, 70.24\% pixel-level accuracy ($\leq100$ px) for initial source detection, and 96.11\% pixel-level accuracy ($\leq100$ px) for point tracking. Our work has established a foundation for AI-assisted bleeding management by enabling prompt surgical intervention through real-time bleeding alerts and bleeding source localization, thereby reducing reliance on repeated water flushing and improving ESD procedural outcomes. Our code and dataset will be available at our project homepage~\url{https://szupc.github.io/ESD_BleedOrigin/}.

\end{abstract}


\end{frontmatter}

\section{Introduction}

\begin{figure*}[!t]
\centering
\includegraphics[width=0.9\linewidth]{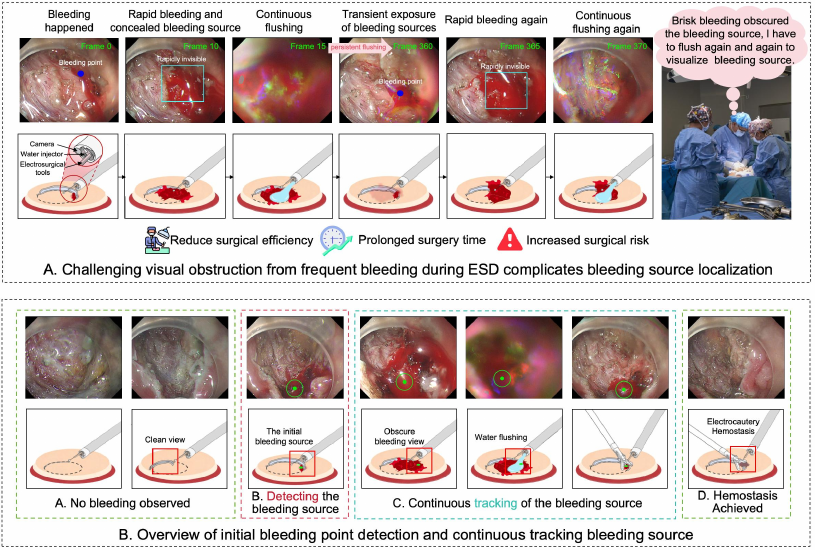}
\caption{Overview of the Motivation and Methodology. (A) Motivation: persistent bleeding obscures the surgical field, necessitating repeated flushing to achieve transient exposure of bleeding sources. This iterative process significantly reduces surgical efficiency, prolongs procedure time, and elevates patient risks, including perforation. A demonstration of this issue is available in our supplementary video (see our  ~\href{https://szupc.github.io/ESD_BleedOrigin/\#motivation_part}{homepage}). (B) The proposed method identifies the initial bleeding source and maintains robust real-time tracking under dynamic visual challenges, ensuring continuous localization until successful hemostasis.}
\label{fig:motivation}
\end{figure*}

Endoscopic Submucosal Dissection (ESD) has revolutionized the treatment of early gastrointestinal neoplasms, enabling precise en-bloc resection while preserving organ function~\cite{saito2014complications}. However, this minimally invasive technique presents unique technical challenges that significantly impact patient safety. Unlike conventional laparoscopic surgery, where cameras and instruments operate independently, ESD employs an integrated endoscopic system where imaging and electrosurgical tools share the same working channel (see Figure~\ref{fig:motivation}A). This configuration creates frequent visual field obstructions, dynamic lighting variations from electrocautery activation, and rapid alternations between clear and blood-obscured views within the confined submucosal space.

Among the various complications associated with ESD, intraoperative bleeding represents the most critical safety concern, occurring in 15-20\% of procedures and directly impacting patient outcomes~\cite{saito2014complications}. The clinical significance of bleeding extends beyond immediate hemorrhage control. Persistent bleeding obscures the surgical field, necessitating repeated water flushing cycles that create only transient visibility windows of mere milliseconds. This iterative process significantly prolongs procedure duration, increases patient exposure to prolonged anesthesia, and elevates risks of serious complications, including perforation, hemodynamic instability, and post-procedural bleeding requiring emergency intervention~\cite{horita2024real}.

Current bleeding management in ESD relies heavily on the surgeon's visual acuity and empirical decision-making. Clinicians depend on direct visual inspection during brief clear intervals, manual irrigation techniques, and subjective assessment of bleeding severity and source location (see Figure~\ref{fig:motivation}A). This conventional approach is inherently limited by human reaction time, visual fatigue during lengthy procedures, and the cognitive burden of maintaining spatial awareness across interrupted visual fields. The challenge is compounded by the transient nature of bleeding manifestations, where hemorrhages may appear as sudden jets, diffuse oozing, or be completely masked by pooling blood and tissue motion.

Artificial Intelligence offers transformative potential for addressing these clinical challenges in ESD bleeding management. AI models can process visual information at speeds far exceeding human perception, potentially identifying subtle bleeding precursors and maintaining precise spatial localization even under adverse conditions~\cite{brzeski2023visual}. This capability enables the development of ``early warning systems'' that could identify hemorrhage before it becomes clinically significant, facilitating proactive rather than reactive intervention. Recent advances in computer vision and deep learning have demonstrated the effectiveness of AI-driven bleeding detection systems in various surgical contexts~\cite{pei2025synergistic,rochmawati2025deep}. For instance, BlooDet~\cite{pei2025synergistic} demonstrates effectiveness in bleeding region and point detection in laparoscopic surgery using dual-task learning approaches. However, the unique visual challenges of ESD procedures present significantly greater complexity than standard laparoscopic environments. The shared endoscope channel configuration creates constant instrument interference, the confined submucosal space amplifies lighting variations, and the frequent water irrigation cycles produce rapid scene transitions that are far more challenging than typical laparoscopic conditions.

The potential benefits of AI-assisted bleeding management in ESD are substantial. Real-time AI detection could reduce physicians' reliance on repeated irrigation, minimize operative delays, and mitigate patient risks such as perforation or excessive fluid absorption~\cite{maulahela2022current}. By providing continuous visual guidance for hemostatic intervention, AI systems could enable surgeons to intervene proactively before bleeding becomes clinically significant, potentially reducing blood loss and improving procedural efficiency~\cite{bamba2022automated}. This paradigm shift from reactive to proactive bleeding management could substantially enhance patient safety and mitigate operative complications.

The development of AI-assisted ESD systems has gained considerable momentum, with researchers exploring applications across multiple procedural aspects. As summarized in Table~\ref{tab:esd_task}, recent advances include topographic mapping for pathological analysis~\cite{xiao2022artificial}, surgical workflow recognition~\cite{cao2023intelligent}, dissection trajectory prediction~\cite{li2023imitation,xu2024etsm}, image enhancement for challenging lighting conditions~\cite{chen2024lightdiff}, tissue segmentation~\cite{xu2025pdzseg}, and multimodal scene understanding~\cite{wang2025endochat}. While these contributions address important aspects of ESD automation, they primarily focus on procedural guidance, image quality improvement, and general scene comprehension.

Despite the critical importance of bleeding management for patient safety, the specific challenge of real-time bleeding source localization, from initial onset detection to continuous spatial tracking, has received limited dedicated attention. Existing bleeding detection research predominantly focuses on broader region segmentation or post-procedural risk prediction~\cite{horita2024real,maulahela2022current,bamba2022automated}, rather than the precise point localization and temporal tracking required for effective hemostatic guidance during live procedures. This represents a significant research gap, particularly given the unique visual challenges of ESD, where instrument interference, water flushing, and blood obscuration create dynamic and unpredictable conditions that are far more complex than those encountered in standard laparoscopic surgery.

The absence of specialized datasets further hinders the development of comprehensive bleeding management systems. While general surgical bleeding datasets exist~\cite{pei2025synergistic,miao2024hemoset}, no large-scale, publicly available benchmark addresses the specific requirements of bleeding source detection and tracking in the ESD environment. This data scarcity limits both algorithm development and standardized evaluation, preventing the establishment of robust performance baselines for this critical surgical AI task.

To address these fundamental gaps, we introduce a comprehensive framework for bleeding source localization in ESD procedures. We present the \textbf{BleedOrigin-Bench}, the first large-scale ESD bleeding source dataset, establishing a standardized benchmark for this essential safety application. Additionally, we propose \textbf{BleedOrigin-Net}, a dual-stage detection-tracking framework that addresses the complete clinical workflow from bleeding onset detection to continuous spatial tracking, providing surgeons with persistent visual guidance for immediate intervention (see Figure~\ref{fig:motivation}B).
To summarize, the key contributions of our work include:

\begin{itemize}
    \item We introduce \textbf{BleedOrigin-Bench}, the first large-scale ESD bleeding source dataset comprising 44 procedures with 106,222 frames and 1,771 precisely annotated bleeding sources, covering 8 anatomical sites and 6 clinical scenarios to establish a standardized benchmark for this critical surgical AI task.

    \item We propose \textbf{BleedOrigin-Net}, the first comprehensive dual-stage framework for bleeding source localization in ESD, featuring BleedOrigin-Detect for initial detection and BleedOrigin-Track for continuous tracking, addressing the complete workflow from bleeding onset detection to continuous spatial tracking.

    \item We develop a \textbf{Multi-Domain Confidence-based Frame Memory (MDCFM)} module that leverages RGB, HSV, and optical flow features to maintain robust temporal context while filtering visual noise, combined with Multi-Domain Gated Attention (MDG) for superior bleeding onset detection.
    
    \item We introduce a \textbf{pseudo-label enhanced tracking strategy} that combines feature matching, trajectory prediction, and Kalman filtering to generate dense supervision from sparse annotations, coupled with parameter-efficient LoRA fine-tuning for robust point tracking under challenging surgical conditions.
    
    \item We establish the first \textbf{standardized evaluation framework} for ESD bleeding source localization with comprehensive comparisons against modern object detection models (YOLOv11/v12), multimodal large language models, and state-of-the-art point tracking methods. Our method achieves state-of-the-art performance with 96.85\% frame-level accuracy ($\pm\leq8$ frames) for bleeding onset detection, 70.24\% pixel-level accuracy ($\leq100$ px) for initial bleeding source detection, and 96.11\% pixel-level accuracy ($\leq100$ px) for point tracking, with practical deployment strategies validated in real surgical scenarios.
\end{itemize}

\begin{table*}[ht]
  \centering
  \caption{Comparison of existing studies in ESD procedures and their relevance to bleeding management. Current approaches focus on workflow recognition, trajectory prediction, and general scene understanding, while our BleedOrigin-Net uniquely addresses the critical unmet need for precise bleeding source detection and continuous spatial tracking during ESD procedures.}
  \resizebox{0.9\linewidth}{!}{
    \begin{tabular}{p{11em}p{15em}p{26em}}
    \toprule
    \textbf{Method} & \textbf{Task Focus} & \textbf{Relevance to ESD Bleeding} \\
    \midrule
    Xiao et al.~\cite{xiao2022artificial}, 2022 & Topographic mapping of resected ESD specimens for pathology & \textit{Irrelevant.} Post-procedural pathological analysis with no connection to intraoperative bleeding detection. \\
    \midrule
    AI-Endo~\cite{cao2023intelligent}, 2023 & Surgical workflow recognition & \textit{Limited relevance.} Recognizes broad surgical phases but lacks precision for real-time bleeding source localization and tracking. \\
    \midrule
    iDiff-IL~\cite{li2023imitation}, 2023  & Dissection trajectory prediction with imitation learning & \textit{Irrelevant.} Focuses on cutting path prediction without addressing bleeding detection or management. \\
    \midrule
    LighTDiff~\cite{chen2024lightdiff}, 2024 & Low-light image enhancement using diffusion models & \textit{Indirect relevance.} Improves image quality but provides no bleeding source detection capabilities. \\
    \midrule
    ETSM~\cite{xu2024etsm}, 2024 & Dissection trajectory suggestion \& confidence map-based safety margin prediction & \textit{Partially relevant.} Aims to prevent bleeding through safer dissection but cannot detect or manage active hemorrhage. \\
    \midrule
    PDZSeg~\cite{xu2025pdzseg}, 2025  & Dissection zone segmentation with visual prompts & \textit{Irrelevant.} Focuses on tissue boundary identification without bleeding source detection capabilities. \\
    \midrule
    EndoChat~\cite{wang2025endochat}, 2025 & Grounded multimodal large language model for surgical scene understanding \& dialogue & \textit{Limited relevance.} General scene understanding without specialized real-time bleeding source localization. \\
    \midrule
    \rowcolor{gray!20}
    BleedOrigin-Net (Ours) & Bleeding source onset detection \& continuous spatial tracking & \textit{Direct solution.} Specifically designed for real-time bleeding onset detection and continuous point tracking in ESD. \\
    \bottomrule
    \end{tabular}%
    }
  \label{tab:esd_task}%
\end{table*}%

\section{Related Work}
The effective management of intraoperative bleeding, particularly in complex procedures such as Endoscopic Submucosal Dissection (ESD), is crucial for ensuring patient safety. While automated detection of bleeding regions has seen progress, the more granular and clinically actionable task of precisely identifying and continuously tracking the specific bleeding source remains significantly underexplored, especially within the demanding visual environment of ESD. This section covers key advancements pertinent to this challenge, including automated bleeding detection, spatiotemporal analysis, hemorrhage tracking, and relevant deep learning strategies.

\subsection{Bleeding Management in Endoscopic Surgery}
Intraoperative bleeding poses substantial risks in minimally invasive surgery, potentially obscuring the surgical field, prolonging procedures, and increasing complication rates~\cite{saito2014complications}. The rapid obscuration of the operative field can hinder the surgical process and elevate the risk of postoperative complications~\cite{pei2025synergistic}. This has motivated extensive research into computational tools for surgical assistance, with computer-aided bleeding detection and localization systems offering substantial clinical value through improved blood loss quantification and enhanced intraoperative decision support~\cite{pei2025synergistic}.

\subsubsection{Bleeding Detection and Region Segmentation}

The ability to identify and delineate bleeding regions has evolved considerably. Early methodologies often relied on traditional image processing, emphasizing color and texture features~\cite{rochmawati2025deep}. These approaches analyzed color spaces like RGB and HSV, using classifiers such as Support Vector Machines (SVMs) to differentiate bleeding from non-bleeding pixels~\cite{brzeski2023visual}. For instance, Yuan et al.~\cite{yuan2015automatic} used saliency maps with SVMs for bleeding detection in Wireless Capsule Endoscopy (WCE) images. However, these methods often struggled with the variability of surgical scenes, dynamic lighting, and visual confounders like reddish tissue. 

The advent of deep learning, particularly Convolutional Neural Networks (CNNs), marked a paradigm shift by enabling automatic learning of discriminative features~\cite{rochmawati2025deep}. Numerous studies have employed CNN architectures for semantic segmentation of bleeding regions or used object detection frameworks like Faster R-CNN~\cite{ren2015faster}, YOLO~\cite{redmon2016you}, and RetinaNet~\cite{lin2017focal} to identify bleeding areas~\cite{garcia2017automatic}. In the Bleeding Alert Map (BAM) framework~\cite{sogabe2023bleeding}, the authors use Generative Adversarial Networks (GANs)~\cite{goodfellow2020generative} for image-to-image translation to identify bleeding areas, notably training on data from mimicking organ systems to overcome data scarcity. More recently, foundation models like the Segment Anything Model (SAM)~\cite{kirillov2023segment} are being adapted for complex medical image segmentation. Hemo-FS-SAM2~\cite{wang2025hemo} proposes to fine-tune SAM2 under few-shot hemorrhage segmentation settings. Besides, BlooDet~\cite{pei2025synergistic} uses SAM 2 for synergetic bleeding region and point detection, underscoring the benefit of leveraging large, pre-trained models for enhanced accuracy. Despite these advancements, challenges remain in handling visual confounders like smoke and water flushing, and the diverse visual manifestations of bleeding.

\subsubsection{Bleeding Source Localization and Point Detection}

While segmenting bleeding areas is valuable, the precise localization of the bleeding source or point is paramount for effective hemostasis. This task is significantly more difficult due to the small, transient, or obscured nature of bleeding sources~\cite{brzeski2023visual}. Hua et al.~\cite{hua2022automatic} proposed a spatiotemporal hybrid model combining RGB data with optical flow to capture motion cues for bleeding source detection. The BAM framework~\cite{sogabe2023bleeding} also targets precise bleeding origin detection using image-to-image translation and trajectory detection algorithms. Recent research has emphasized synergistic, dual-task learning for concurrently detecting bleeding regions and localizing points. The BlooDet framework~\cite{pei2025synergistic}, for example, employs a dual-branch design where the mask (region) and point localization branches interact and guide each other. These efforts highlight a clear trend: moving beyond general region detection to provide the actionable, point-specific information endoscopy doctors need for targeted intervention, a gap that is particularly evident in the complex field of ESD.

\subsubsection{Intelligent Systems for Active Hemostasis Support}

Beyond passive detection, research is increasingly focused on intelligent systems that actively assist endoscopy doctors. This reflects a progression from identifying problems to suggesting or performing corrective actions. For example, Richter et al.~\cite{richter2021autonomous} developed an autonomous robotic suction system that uses image-based blood flow detection to clear the surgical field, preparing the site for hemostasis. Other work has explored autonomous suction in simulated laparoscopic scenes~\cite{ou2025learning}. Alert systems like BAM~\cite{sogabe2023bleeding} provide real-time visual cues to direct endoscopy doctors' attention to bleeding sources. Furthermore, AI systems are being developed to identify specific vessels requiring post-ESD coagulation (PEC) to prevent delayed bleeding, a critical aspect of hemostasis management~\cite{fujinami2024development}. These systems aim to reduce the surgeon's cognitive load and improve response times, ultimately enhancing surgical safety.

\subsection{Advanced Video Analysis for Dynamic Surgical Environments}

Surgical procedures are inherently dynamic, requiring robust video analysis techniques that can interpret temporal evolution and motion to provide effective AI-driven assistance.

\subsubsection{Temporal Event Localization: Identifying Bleeding Onset}

Precisely identifying the start time, or onset, of critical events, such as bleeding, is crucial for timely intervention. This is challenging as bleeding is often a progressive phenomenon, not an instantaneous change. Modern approaches for temporal event localization increasingly leverage Transformer architectures to capture long-range dependencies~\cite{muksimova2025cross}. Frameworks like TEMPURA~\cite{cheng2025tempura} use masked event prediction for fine-grained temporal grounding. Memory-augmented networks, such as the Memory-Augmented Transformer (MATR)~\cite{song2024online}, are also highly effective; as they utilize a memory queue to preserve past features, allowing them to leverage long-term context for inference. This is particularly relevant for bleeding onset detection, where recognizing the subtle transition from a ``clean view'' to a bleeding state requires comparing the current scene with a recent, non-bleeding baseline.

\subsubsection{Visual Tracking for Continuous Hemorrhage Monitoring}

Continuously and accurately tracking key points, such as a bleeding source, is essential for maintaining situational awareness, especially when visibility is intermittent. Current state-of-the-art visual point trackers are predominantly Transformer-based. Prominent examples include TAPIR~\cite{doersch2023tapir} and BootsTAP~\cite{doersch2024bootstap}, which use a two-stage matching and refinement methodology; CoTracker~\cite{karaev2024cotracker}, which jointly tracks numerous points to leverage their dependencies and improve robustness during occlusion; and Track-On~\cite{aydemir2025track}, an online framework that uses memory modules to facilitate long-term tracking. However, a significant domain gap exists between the general or synthetic datasets on which these models are trained and the unique challenges of surgical video. Surgical scenes feature highly deformable tissues, frequent occlusions from tools and fluids, and camera jitter, which can degrade the performance of generic trackers~\cite{you2024pace}. In point tracking training framework, it is usually necessary to sample more points per frame (about 512 to 1024 points) for normal training or fine-tuning. This necessitates specialized model adaptation strategies to make these powerful trackers effective in a clinical setting.

\subsection{Resource-efficient Learning in Surgical Data Analysis}
The scarcity of annotated surgical data poses a significant challenge for developing robust AI models. To address this, several strategies have emerged, focusing on leveraging pre-trained models and semi-supervised learning techniques.

\subsubsection{Addressing Data Scarcity: Semi-Supervised Learning (SSL) and Pseudo-Labeling}

The high cost and expertise required for annotating medical videos have made SSL a critical area of research~\cite{li2023semi,tanwani2020motion2vec,shi2021semi,li2025semivt}. SSL leverages abundant unlabeled data alongside limited labeled data to improve model performance. Pseudo-labeling is a prominent SSL technique where a model's own confident predictions on unlabeled data are used as new training labels. This approach has been successfully applied in surgical AI for tasks like phase recognition~\cite{shi2021semi,li2025semivt}, often incorporating temporal consistency regularization to ensure stable predictions.

\subsubsection{Adapting Large Foundation Models: Parameter-Efficient Fine-Tuning (PEFT)}

While the emergence of large foundation models has brought transformative capabilities to AI, their substantial size poses computational challenges for full fine-tuning, especially in specialized domains like surgery, where data availability is often constrained~\cite{dutt2023parameter,liang2025vision,noh2025narrative}. Parameter-Efficient Fine-Tuning (PEFT) techniques offer a solution by enabling adaptation of these models through the modification of only a small subset of their parameters~\cite{han2024parameter}. A notable PEFT method is Low-Rank Adaptation (LoRA)~\cite{hu2022lora}, which introduces small, trainable low-rank matrices into the architecture of a frozen pre-trained model. This allows for efficient model specialization by adapting these injected matrices without altering the extensive original weights~\cite{qu2025adapting}. LoRA and similar PEFT approaches are particularly advantageous for medical imaging, facilitating the stable and effective adaptation of powerful general-purpose models to nuanced surgical tasks, even when faced with low-data regimes.

\subsection{Bleeding Datasets and Comparison}

Progress in surgical AI is fundamentally dependent on the availability of high-quality, annotated datasets~\cite{nie2024review}. While general benchmarks exist, specialized datasets are crucial for addressing specific clinical problems. 
For bleeding detection and related tasks, a number of datasets have been introduced, each with its specific focus and scope. Table~\ref{tab:dataset_comp} provides a comparative overview of existing surgical bleeding datasets, highlighting their respective characteristics, including the year of introduction, targeted tasks, data types, annotation types, and the volume of data in terms of videos/clips and frames/images.
As illustrated in the table, existing resources such as SurgBlood~\cite{pei2025synergistic} offer annotations for bleeding regions and points in laparoscopic surgery, HemoSet~\cite{miao2024hemoset} concentrates on blood segmentation in robotic surgery, and datasets from capsule endoscopy like WCEBleedGen~\cite{handa2024wcebleedgen} provide annotations for bleeding frame classification, region detection, and segmentation. Other notable contributions include early work on bleeding frame and source detection by Rahbar et al.~\cite{rahbar2020entropy}, blood region and flow detection by Richter et al.~\cite{richter2021autonomous}, and bleeding source detection in laparoscopy by Hua et al.~\cite{hua2022automatic}.

Despite these efforts, a significant void persists: there is a lack of a comprehensive, large-scale, publicly accessible dataset specifically addressing the nuanced requirements of bleeding source detection and continuous tracking within the uniquely demanding visual and operational context of Endoscopic Submucosal Dissection (ESD). Bridging this gap by establishing such a benchmark resource is crucial for fostering the development and rigorous evaluation of AI-driven solutions tailored to this critical aspect of ESD.

\begin{table*}[!t]
  \centering
  \caption{Comprehensive comparison of existing surgical bleeding datasets with our proposed BleedOrigin-Bench about the ESD bleeding source dataset. Our dataset uniquely addresses bleeding source detection and tracking in ESD procedures with the largest scale of annotated bleeding sources (1,771) across 485 video clips, filling a critical gap in ESD-specific bleeding management research.}
  \resizebox{0.9\linewidth}{!}{
    \begin{tabular}{lllllll}
    \toprule
    \textbf{Dataset Name} & \multicolumn{1}{l}{\textbf{Year}} & \multicolumn{1}{l}{\textbf{Tasks}} & \multicolumn{1}{l}{\textbf{Data Type}} & \multicolumn{1}{l}{\textbf{Annotation Type}} & \multicolumn{1}{l}{\textbf{Videos/Clips}} & \multicolumn{1}{l}{\textbf{Frames/Images}} \\
    \midrule
    Rahbar et al.~\cite{rahbar2020entropy} & 2020  & \multicolumn{1}{p{12em}}{Bleeding frame occurrence and source detection} & \multicolumn{1}{l}{Laparoscopic surgery} & \multicolumn{1}{p{8.28em}}{Bleeding frame and source (coordinates)} & \multicolumn{1}{l}{15 videos} & \multicolumn{1}{l}{Not specified} \\
    Richter et al.~\cite{richter2021autonomous} & 2021  & \multicolumn{1}{p{12em}}{Blood region and flow detection} & \multicolumn{1}{p{9.5em}}{Simulated scenes, Real-life trauma (thyroidectomy)} & \multicolumn{1}{p{8.28em}}{Blood region mask} & \multicolumn{1}{p{7.22em}}{6 simulated scenes, 1 in vivo thyroidectomy video} & \multicolumn{1}{p{7.055em}}{366 simulated frame} \\
    Hua et al.~\cite{hua2022automatic} & 2022  & \multicolumn{1}{l}{bleeding source detection} & \multicolumn{1}{l}{Laparoscopic Surgery} & \multicolumn{1}{p{8.28em}}{bleeding sources (center of marked box)} & \multicolumn{1}{p{7.22em}}{12 clips from 10 surgeries} & \multicolumn{1}{p{7.055em}}{2,665 images (1,339 w/ bleeding)} \\
    Rabbani et al.~\cite{rabbani2022video} & 2022  & \multicolumn{1}{l}{Blood Segmentation} & \multicolumn{1}{p{9.5em}}{Gynecologic laparoscopic surgeries} & \multicolumn{1}{l}{Binary blood mask} & \multicolumn{1}{l}{96 videos} & \multicolumn{1}{l}{Not specified} \\
    BAM~\cite{sogabe2023bleeding}   & 2023  & \multicolumn{1}{l}{Bleeding source estimation} & \multicolumn{1}{l}{Mimicking organ} & \multicolumn{1}{p{8.28em}}{Bleeding source coordinates and circular alert map} & \multicolumn{1}{l}{Not specified} & \multicolumn{1}{l}{3735 frames} \\
    Ou et al.~\cite{ou2024autonomous} & 2024  & \multicolumn{1}{p{12em}}{Autonomous suction of simulated blood} & \multicolumn{1}{p{9.5em}}{Simulated laparoscopic scenes} & \multicolumn{1}{l}{Binary blood mask} & \multicolumn{1}{l}{Not specified} & \multicolumn{1}{l}{Not specified} \\
    HemoSet~\cite{miao2024hemoset} & 2024  & \multicolumn{1}{l}{Blood Segmentation} & \multicolumn{1}{p{9.5em}}{Live animal robotic surgery} & \multicolumn{1}{l}{Binary blood mask} & \multicolumn{1}{l}{11 videos} & \multicolumn{1}{p{7.055em}}{102616 frames (w/ 857 labeled)} \\
    WCEBleedGen~\cite{handa2024wcebleedgen} & 2024  & \multicolumn{1}{p{12em}}{Bleeding classification, detection and segmentation} & \multicolumn{1}{p{9.5em}}{Wireless capsule endoscopy (WCE)} & \multicolumn{1}{p{8.28em}}{Binary bleeding/non-bleeding label, blood region (bounding box and mask)} & \multicolumn{1}{l}{Not specified} & \multicolumn{1}{p{7.055em}}{2618 frames (1309 w/ bleeding)} \\
    SurgBlood~\cite{pei2025synergistic} & 2025  & \multicolumn{1}{p{12em}}{Bleeding region and point detection} & \multicolumn{1}{p{9.5em}}{Laparoscopic Cholecystectomy Surgery } & \multicolumn{1}{p{8.28em}}{Bleeding regions (pixel-level), bleeding sources (coordinates)} & \multicolumn{1}{p{7.22em}}{95 surgical video clips} & \multicolumn{1}{l}{5,330 frames} \\
    \rowcolor{gray!20}
    BleedOrigin-Bench (Ours)   & 2025  & \multicolumn{1}{p{12em}}{Bleeding onset detection and continuous bleeding source tracking} & \multicolumn{1}{p{9.5em}}{Endoscopic Submucosal Dissection (ESD)} & \multicolumn{1}{p{9em}}{bleeding sources (coordinates), temporal onset frames} & \multicolumn{1}{p{8em}}{485 video clips from 44 procedures} & \multicolumn{1}{p{7.8em}}{106,222 frames (41,526 w/ bleeding sources)} \\
    \bottomrule
    \end{tabular}%
    }
  \label{tab:dataset_comp}%
\end{table*}%

\subsection{Summary of Gaps and Motivation for Present Study}

The comprehensive review of existing literature reveals a critical unmet need in the context of Endoscopic Submucosal Dissection (ESD). While advancements have been made in automated bleeding detection and, to some extent, localization, a significant gap persists. Current methodologies often fall short of providing a holistic solution that progresses from the macro-level identification of a bleeding event down to the micro-level precision required for effective intervention in ESD. This involves a hierarchical approach: first, identifying the \textbf{bleeding frame} to pinpoint the temporal onset of hemorrhage; second, detecting the broader \textbf{bleeding region} to understand its extent; and critically, locating the precise \textbf{bleeding source coordinate} for targeted therapy.

Despite the progress in individual aspects, there is a distinct lack of systems capable of robustly performing precise, real-time detection of initial bleeding sources and subsequently maintaining continuous, accurate tracking of these points, especially within the uniquely challenging visual and dynamic environment of ESD. Many existing approaches are tailored to less complex surgical settings, focus predominantly on broader segmentation of the bleeding region, or lack the integrated spatio-temporal coherence essential for reliable guidance during ESD, a procedure characterized by its high risk of bleeding, frequent instrument-tissue interactions, and difficult visualization due to motion of smoke, fluid, and tissue.

Therefore, the primary motivation for the research detailed in this paper is the development of a comprehensive, dual-stage system that specifically addresses these multifaceted challenges in ESD. Our work aims to bridge the identified gaps by integrating nuanced bleeding onset detection, pinpoint spatial localization of the hemorrhage source, and robust tracking capabilities to withstand occlusions and dynamic scene changes, thereby providing reliable, continuous guidance for endoscopists.

\begin{figure*}[!ht]
\centering
\includegraphics[width=0.9\linewidth]{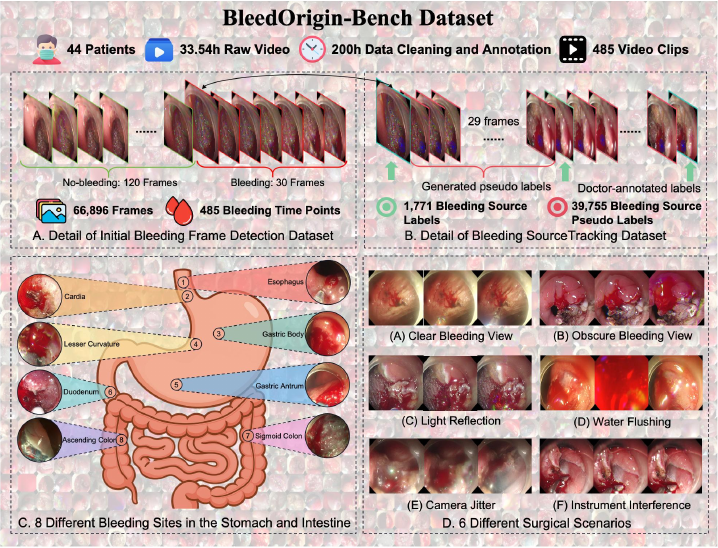}
\caption{Overview of our BleedOrigin-Bench dataset. We selected $485$ bleeding video clips from $44$ patients for analysis, processed in two stages: (1) initial bleeding frame and bleeding source detection, and (2) subsequent bleeding source tracking. The resulting datasets include: A. Detection set: $66,896$ frames with $485$ bleeding time points; B. Tracking set: 1,771 frames with manually annotated bleeding sources and an additional 39,755 frames augmented with bleeding source pseudo-labels; C. The dataset features anatomical diversity: 8 sites (gastric antrum, duodenum, etc.); D. clinical challenges: 6 scenarios (obscured bleeding views, camera jitter, water flushing, etc.)}
\label{fig:datase_overview}
\end{figure*}

\section{BleedOrigin-Bench Dataset}

To address the critical knowledge gap in AI-assisted bleeding management during Endoscopic Submucosal Dissection (ESD), we introduce a purpose-built ESD surgical video dataset that prioritizes high-risk bleeding scenarios.

\subsection{Dataset Collection and Processing}

This study utilize a retrospective dataset comprising $44$ endoscopic submucosal dissection (ESD) procedures, acquired from Qilu Hospital, Shandong University, with written patient consent for research use. The study is approved by the Institutional Ethics Committee (Approval No.DWLL-2021-021). All data are fully anonymized, retaining only imaging system outputs for analysis. The study protocol adhered to institutional ethical guidelines and received approval from the relevant ethics review boards at all participating sites.
The $44$ surgical videos are captured at $30$ frames per second (FPS) with the resolution of ${1920 \times 1080}$ pixels and ${1280 \times 720}$ pixels. Individual video durations range from tens of minutes to 1–2 hours (mean: 46.14 minutes), totaling $33.54$ hours of footage (see Figure~\ref{fig:datase_overview}). All videos are uniformly downsampled to 1 FPS. Since the videos are directly captured from surgical recording devices, we manually crop out irrelevant regions such as patient monitoring parameters, interface overlays, and black margins. Only the endoscopic view is retained to ensure data quality with the resolution of ${1240 \times 1080}$ and ${780 \times 670}$.

\begin{figure*}[t]
\centering
\includegraphics[width=0.9\linewidth]{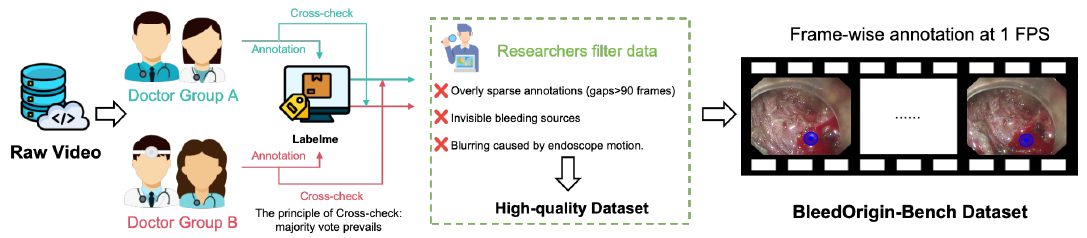}
\caption{Overview of the dataset annotation pipeline. Our multi-stage annotation pipeline is used to create the high-quality BleedOrigin-Bench dataset. Raw videos are independently annotated by two groups of clinicians. A majority vote principle is used for cross-checking, followed by a researcher-led filtering step to remove ambiguous or low-quality labels, resulting in a curated dataset with precise frame-wise annotations.}
\label{fig:annotation_pipeline}
\end{figure*}

\subsection{Bleeding Source Annotation}

Given the dataset's scale, diversity, and time-intensive annotation requirements, particularly, due to high inter-frame similarity in consecutive frames, we systematically sample frames at $30$ fps from $44$ endoscopic procedures, obtaining $106,222$ high-resolution images to build the dataset for the initial bleeding source detection and tracking tasks. Among these, $1,771$ frames have bleeding sources annotated at 1 FPS by four board-certified endoscopists using the LabelMe tool\footnote{\url{https://github.com/wkentaro/labelme}} (see Figure~\ref{fig:annotation_pipeline}). The BleedOrigin-Bench dataset is created through our rigorous multi-stage annotation pipeline: Independent annotation by two separate clinician teams; Consensus validation using majority voting; Expert refinement with researcher-led filtering to eliminate ambiguous or low-quality labels. This process ensures high-quality, precise, frame-wise annotations for all samples. In addition, clinical experts annotate 3 full-length videos, each exceeding 15 seconds in duration and sampled at 30 FPS. These videos are densely annotated with bleeding source locations on every frame, capturing the complete progression from clean field to the onset and continuation of bleeding. This densely annotated subset is not included in the main test set, but rather serves as a separate benchmark for evaluating the model's ability to detect and continuously track bleeding sources throughout the entire bleeding process.

\subsection{BleedOrigin-Bench Dataset Statistics}
From the collection, $485$ bleeding video clips are selected for detection and tracking analysis. Each clip is processed in two stages: (1) bleeding source detection in the initial segment, followed by (2) bleeding source tracking in subsequent frames. This yields two datasets: (i) An initial bleeding frame and bleeding source detection dataset containing $66,896$ frames with $485$ bleeding time points (see Figure~\ref{fig:datase_overview}A); (ii) A bleeding source tracking dataset comprising $1,771$ manually annotated frames and $39,755$ pseudo-labeled frames generated by our method (Figure~\ref{fig:datase_overview}B). The dataset encompasses bleeding events across diverse anatomical locations, including the gastric antrum, the gastric body, duodenum, ascending colon, esophagus, sigmoid colon, cardia, and lesser curvature, ensuring comprehensive coverage of bleeding site diversity in ESD procedures (see Figure~\ref{fig:datase_overview}C). In addition, the dataset covers six clinically challenging scenarios: clear bleeding view, obscure bleeding view, camera jitter, light reflection, water flushing, and instrument interference (see Figure~\ref{fig:datase_overview}D)

\subsubsection{Details of the Bleeding Source Detection Dataset}
\begin{figure*}[t]
\centering
\includegraphics[width=0.9\linewidth]{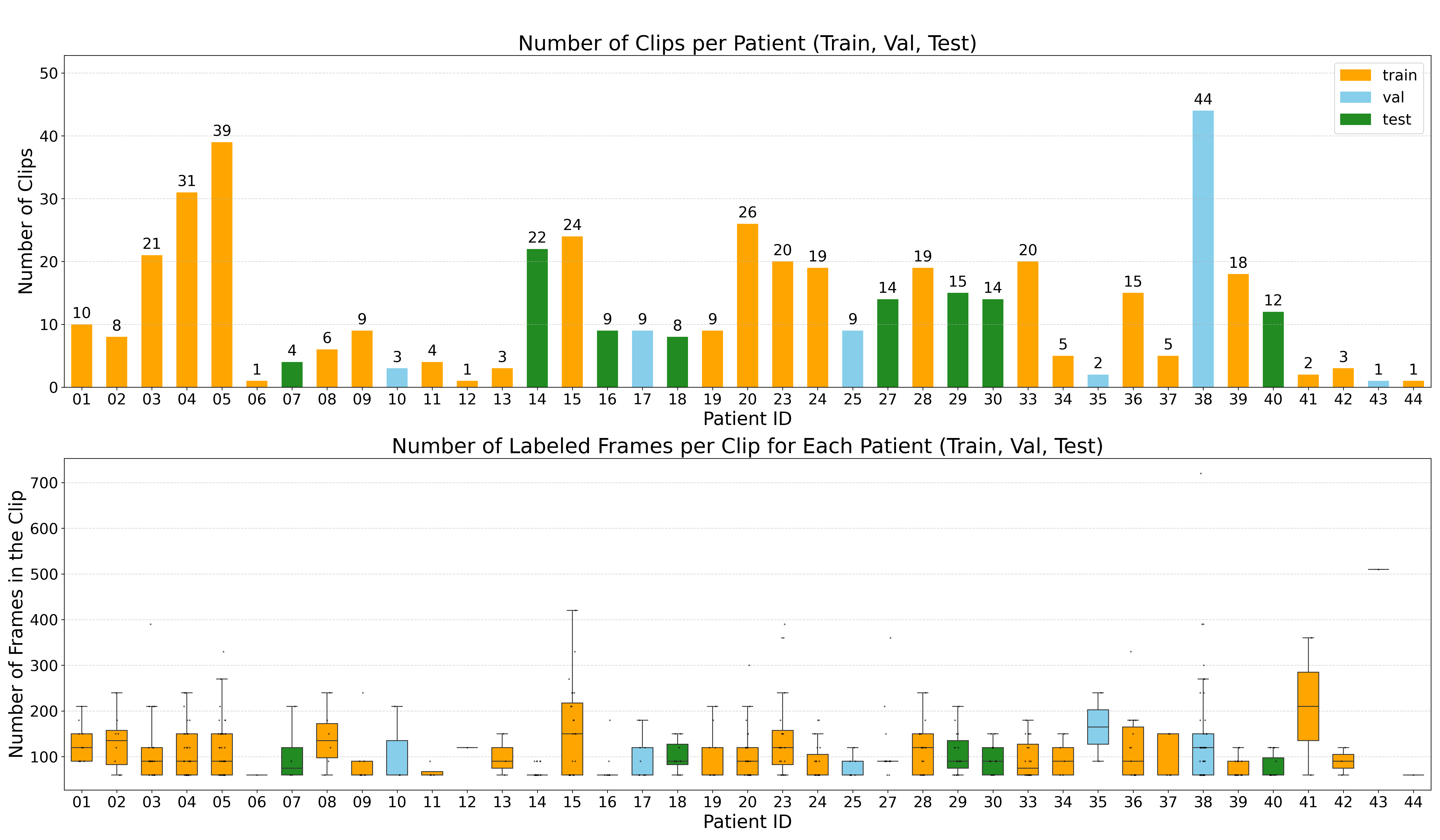}
\caption{Dataset statistics by patient. (Top) Distribution of video clips per patient across train(yellow), validation(blue), and test(green) splits. (Bottom) The number of annotated frames per clip is shown as box plots for each patient.}
\label{fig:detection_dataset}
\end{figure*}

For the initial bleeding source detection dataset, data from a total of \textbf{44 patients} are included. Due to the variability of surgical procedures, the number of video sub-segments (referred to as ``clips'') varied across patients, with the specific clip distribution illustrated in Figure~\ref{fig:detection_dataset}. The dataset is split into training, validation, and test sets in a 4:1:1 ratio at the patient level to ensure that each patient's data appears exclusively in one subset, preserving experimental independence and avoiding data leakage. In total, \textbf{485 video clips} are extracted. The training set consists of 25 patients and 319 clips, the validation set comprises 6 patients and 68 clips, and the test set consists of 8 patients and 98 clips. Each clip consists of 150 consecutive frames, and the initial bleeding source consistently occurs at frame 120. Accordingly, only frame 120 is annotated with the ground truth coordinates of the bleeding source.

During training, to mitigate overfitting and enhance model robustness, we randomly discard the first 0 to 60 frames of each video sub-segment. Specifically, the model begins reading from any frame between \( I_0 \) and \( I_{60} \), thus introducing variability in the training inputs. This random frame-skipping strategy is only applied during the training phase. For validation and testing, the full clips are used without frame skipping to ensure consistent and reliable evaluation.

\subsubsection{Details of the Bleeding Source Tracking Dataset}
For the bleeding source tracking dataset in ESD procedures, each clip contains a complete video stream along with annotations sampled at 1 FPS (one labeled frame every 30 frames). Detailed annotation distributions are illustrated in Figure~\ref{fig:detection_dataset}. The dataset split strategy is consistent with that used in the detection task, maintaining the same number of clips and ensuring comparable experimental settings.

Furthermore, since existing tracking datasets such as MOVi-E or MOVi-F~\cite{greff2021kubric} are composed of short clips with a fixed length of 25 frames, we introduce two definitions to adapt our dataset accordingly. Specifically, we define a \textbf{Short Clip} as a subset within a clip consisting of 31 consecutive frames, where both the first and the last frames are annotated with the coordinates of the bleeding source. In contrast, a \textbf{Long Clip} refers to the entire video clip within a given clip, with a length ranging from 31 to 731 frames. In the long clip setting, the bleeding source is annotated every 30 frames with its corresponding coordinates.

\subsubsection{Dataset Splitting Strategy}
We conducted both the initial bleeding source detection and continuous bleeding source tracking experiments using the same patient-level and clip-level splits. 
For each video clip, detection is performed on the initial segment to identify the bleeding source, while tracking is executed on the subsequent segment to monitor the point trajectory. To prevent data leakage and ensure robust evaluation, the dataset is partitioned at the patient level into training, validation, and test sets in a 4:1:1 ratio, maintaining strict separation between all subsets. The detailed distribution and counts are shown in Figure~\ref{fig:detection_dataset}.

\section{Methodology}
Having established the BleedOrigin-Bench dataset, we now present our computational framework for addressing the critical challenge of bleeding source localization in ESD procedures.
We formulate bleeding source localization as a sequential two-stage process: (1) \textbf{Detection}: identifying the initial bleeding frame and localizing the bleeding source; and (2) \textbf{Tracking}: continuously monitoring the bleeding source across subsequent frames.

ESD procedures present unique visual challenges, including instrument occlusion, water flushing, camera jitter, and blood obscuring the surgical field, making frame-by-frame detection unreliable. To address these challenges, we propose the \textbf{BleedOrigin-Net}, a dual-stage framework that integrates \textbf{BleedOrigin-Detect} for robust bleeding onset detection with the \textbf{BleedOrigin-Track} for continuous spatial tracking. The detection component ensures high precision in identifying the onset of bleeding, while the tracking module maintains robust temporal consistency under dynamic conditions. Each component can be independently optimized. Additionally, the tracking module can operate independently when clinicians provide manual initialization, offering flexible deployment for different clinical scenarios.

During deployment, the two independently trained modules execute sequentially within a unified pipeline, enabling seamless transition from initial bleeding detection to continuous point tracking throughout the surgical procedure.

\begin{figure*}[t]
\centering
\includegraphics[width=0.9\linewidth]{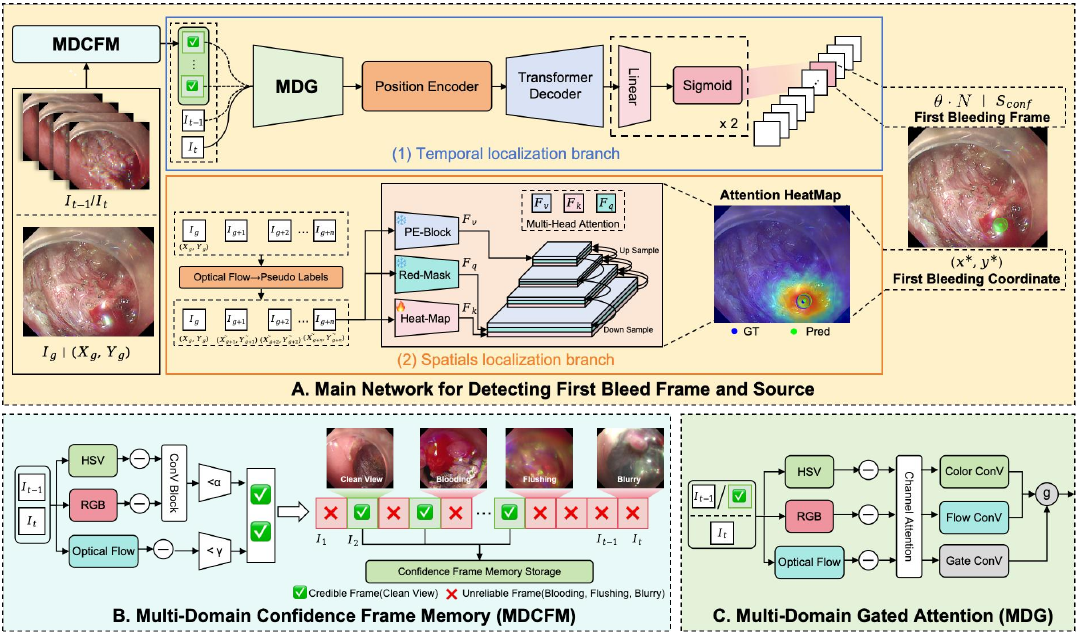}
\caption{Overview of the proposed BleedOrigin-Detect for detecting the initial bleeding frame and point in ESD procedures. \textbf{A} The main network architecture consists of a vision backbone, a position encoder, and a transformer decoder that jointly predict the initial bleeding frame and its corresponding point. \textbf{B} The Multi-Domain Confidence-based Frame Memory (MDCFM) module selectively retains temporally consistent frames based on cross-domain cues (RGB, HSV, optical flow) to suppress occlusions and noise. \textbf{C} The Multi-Domain Gated Attention (MDG) module adaptively fuses multi-domain representations to guide spatial feature encoding in challenging bleeding scenarios.
}
\label{fig:detection}
\end{figure*}

\subsection{BleedOrigin-Detect: Detecting the Initial Bleeding Frame and the Initial Bleeding Source}
It is critically important to detect the precise frame in which bleeding first occurs and to localize the exact spatial coordinates of the bleeding site, as accurate localization directly influences the effectiveness of subsequent bleeding source tracking. However, this task faces several significant challenges. As shown in Figure~\ref{fig:motivation}A, ESD employs a single endoscope that integrates both imaging and electrosurgical functions within the same working channel. This integration creates constant visual field changes as the tool and the camera share the same working channel. Additionally, diffuse bleeding severely compromises visibility, making precise detection of bleeding onset frames and accurate spatial localization of the bleeding source critically challenging yet essential for effective hemostatic intervention. To address these challenges, our complete ESD initial bleeding frame and point detection method BleedOrigin-Detect, as shown in Figure~\ref{fig:detection}A, decompose the bleeding detection task into two key components: (1) Temporal Localization: identifying the initial bleeding frame; and (2) Spatial Localization: pinpointing the initial bleeding source in the initial bleeding frame. 

\subsubsection{Temporal Localization: Detecting the Initial Bleeding Frame} 

Detecting the initial bleeding frame is a challenging task, as bleeding is a temporally progressive event rather than an instantaneous change. It typically begins with a clean surgical field and gradually evolves as blood diffuses from a focal point. This temporal characteristic makes it difficult to determine the onset of bleeding using single-frame classification alone. It is crucial to emphasize that our task focuses on detecting the first frame when bleeding occurs (bleeding onset detection), which is significantly more challenging than simply identifying any frame containing bleeding. While general bleeding frame detection can rely on obvious visual cues such as visible blood pools or diffuse hemorrhage, initial bleeding frame detection requires capturing the subtle transition moment from a clean field to the initial emergence of bleeding, demanding precise temporal localization and heightened sensitivity to early bleeding indicators.

To address this, it is crucial to incorporate information from preceding non-bleeding frames as temporal references, which can help the model better recognize the subtle transition into bleeding. Memory-based modules, widely adopted in video object detection and tracking~\cite{xiao2018video, liu2019looking, yang2024samurai}, offer a natural solution for modeling such temporal dependencies by storing and recalling contextual features.  However, real-world surgical scenes are often disrupted by visual disturbances such as motion blur, camera jitter, and instrument interference, which can introduce visual noise and corrupted memory representations, potentially degrading the model's prediction performance and temporal consistency. 
Therefore, we propose a memory mechanism that enables our memory storage to maintain the recent historical frame $I_{t-1}$ and selectively retains clean view frames from the procedural history, providing both short-term temporal continuity and long-term reference baselines for bleeding onset detection. The ``clean view'' frames encompass frames that are both blood-free and have unobstructed, clearly visible anatomical structures,

To select ``clean view'' frames as keyframes for building the memory under the challenges posed by visual disturbances, we propose the \textbf{Multi-Domain Confidence-based Frame Memory Module} (MDCFM), as illustrated in Figure~\ref{fig:detection}B. For each pair of consecutive input frames $I_{t-1}$ and $I_{t}$, we compute inter-frame differences from multiple perceptual domains, including color-based features (RGB and HSV) and motion-based features (optical flow), to capture both appearance and dynamic changes. For the RGB and HSV modalities, denoted as $f_{\text{RGB-HSV}}(\cdot)$, frame-wise differences are first computed and subsequently passed through a stack of three convolutional layers for feature extraction. In parallel, optical flow representations, expressed as $f_{\text{Flow}}(\cdot)$, are extracted to capture motion-related changes. The outputs from both branches are concatenated and processed by a fully connected layer, followed by a Sigmoid activation, yielding normalized confidence scores (ranging from 0 to 1) for each perceptual domain. Only when both confidence values fall below the respective and predefined thresholds ($\alpha=0.5$, $\gamma=0.5$) do we consider the previous frame $I_{t-1}$ to be ``Clean View'' and store it in the \textbf{Confidence-based Frame Memory Storage} (CFMS). The selection of keyframes $I_{\text{k}}$ is formally defined is shown in Equation \eqref{equation: keyframe}:

\begin{equation}
\label{equation: keyframe}
I_{\text{k}} = 
\begin{cases}
I_{t-1}, & \text{if } f_{\text{RGB-HSV}}(I_{t-1}, I_t) < \alpha \;\land\; f_{\text{Flow}}(I_{t-1}, I_t) < \gamma \\
\varnothing, & \text{else}.
\end{cases}
\end{equation}

Given two consecutive frames, $I_{t-1}$ and $I_t$, we compute feature differences not only between $I_t$ and $I_{t-1}$, but also between $I_t$ and each previous frame stored in the CFMS. We introduce the \textbf{Multi-Domain Gated Attention} (MDG), as shown in Figure~\ref{fig:detection}C. For each comparison, we extract features from three complementary perceptual domains: RGB, HSV, and Optical Flow, resulting in domain-specific difference maps $F_{\mathrm{RGB}}$, $F_{\mathrm{HSV}}$, and $F_{\mathrm{Flow}}$ $\in \mathbb{R}^{C \times H \times W}$. These features are then fused through a channel attention, which adaptively balances the contribution of each domain by reweighting channel-wise responses, thereby enhancing the model’s ability to focus on discriminative cues across color and motion. The resulting fused feature $F_{c}$ $\in \mathbb{R}^{3C \times H \times W}$ is subsequently passed through three parallel convolutional branches with identical architectures to extract domain-specific representations for color and motion. The gated attention is then employed to adaptively fuse the outputs from the three branches, allowing the model to emphasize the informative domains based on contextual cues.

Due to the lack of temporal context, it is challenging to accurately classify the bleeding onset time using only a single frame or a pair of adjacent frames. To address this limitation, we employ a sliding window approach with a fixed size of $N = 60$ frames, which advances consecutively across the video clip to capture temporal dynamics. For each frame $I_t$ in the clip, visual features are first extracted by MDG Module. These features are then augmented with positional encodings to retain temporal ordering information and subsequently fed into a standard Transformer architecture. The temporal modeling module consists of four stacked Transformer encoder layers. Each encoder layer contains a multi-head self-attention mechanism with eight heads, followed by a feed-forward network. The output of the Transformer is passed through two identical fully connected layers to adjust feature dimensionality. Finally, a sigmoid activation function is applied to generate two outputs: (1) the normalized temporal location $\theta \in [0, 1]$ of the initial bleeding frame within the input clip; (2) the corresponding confidence score $S_{\text{conf}}^{(t)}\in [0, 1]$, indicating the reliability of the predicted onset time. The complete process is shown in Equation \eqref{equation: bleed_frame}:
\begin{equation}
\label{equation: bleed_frame}
\resizebox{0.9\columnwidth}{!}{$
{t}_{\text{bleed}} = \min_{t \in \mathcal{T}} \left\{\, t + \left\lfloor \theta^{(t)} \cdot N \right\rfloor\ \middle|\ (\theta^{(t)},\ S_{\text{conf}}^{(t)}) = f_{bf}\left( V_{t:t+N} \right),\ S_{\text{conf}}^{(t)} > 0.5 \,\right\}.
$}
\end{equation}

We set the full length of the complete video to be $\mathcal{T}$. The predicted bleeding onset frame ${t}_{bleed}$ is determined by applying a sliding window approach over the complete video $V$. Each window of length $N$ is processed by the initial bleeding frame prediction model $f_{bf}(\cdot)$, which outputs a normalized onset position $\theta^{(t)} \in [0, 1]$. Only predictions with $S_{\text{conf}}^{(t)} > 0.5$ are considered valid. Among these, the earliest estimated frame index, computed as $t + \lfloor \theta^{(t)} \cdot N \rfloor$, is selected as the final onset frame $I_{{t_{b}}}$.

\subsubsection{Spatial Localization: Detecting the Initial bleeding source}

After identifying the bleeding source, localizing it based on a single frame remains a particularly challenging task. Conventional RGB color-based detection methods often result in erroneous predictions due to the complexity and variability of endoscopic visual scenes~\cite{hua2022automatic, liu2022research}. To address this, we propose a multi-strategy approach to enhance robustness and accuracy.

First, we generate pseudo-labels to introduce supervision across multiple temporal frames. Specifically, given a manually annotated bleeding frame $I_g$ and the corresponding bleeding coordinates $(X_{g}, Y_{g})$, we employ optical flow estimation to approximate the location $\tilde{(X_{g+n}}, \tilde{Y_{g+n})}$ of the bleeding source in a subsequent frame $I_{g+n}$, where $n = 10$. This strategy provides rough annotations for future frames, enhancing the temporal continuity in training supervision. Importantly, during the training phase, we do not use the initially predicted bleeding frame for detection. Instead, the ground-truth bleeding frame $I_g$ is directly used as input. This design choice accelerates model convergence and stabilizes learning by eliminating early-stage prediction noise.

To extract discriminative features from a frame, we adopt a multi-branch feature encoding strategy. The \textbf{Perception Encoder Block (PE-Block)}~\cite{bolya2025PerceptionEncoder} is a pre-trained multimodal feature extraction module designed to capture both color and depth information from endoscopic images, which can achieve better results than some other feature extraction modules, such as DINOv2~\cite{oquab2023dinov2}. There are three versions, PE-Core, PE-Lang and PE-Spatial. Our method selects PE-Spatial, which can better perceive space, and Table~\ref{tab:bleeding_frame_point} also shows the difference between these pre-trained vision encoder blocks. The \textbf{Red-Mask} module is a handcrafted, non-trainable component that generates a red heatmap by mapping the intensity of the R channel in the RGB space, highlighting the bleeding regions, which typically correspond to the saturated red areas. In addition, we employ a \textbf{Heat-Map} module, composed of 2 consecutive $3 \times 3$ Conv layers followed by an upsampling Conv, to generate a dense response map aligned with the input image resolution.

Due to the relatively low spatial resolution ($W \times H$) of feature maps extracted by visual encoders such as the PE-Block or DINOv2, we adopt a Feature Pyramid Network (FPN)-style fusion strategy to enhance multi-scale representation. Specifically, we construct three levels of features: heatmap feature $F_h$ and red-mask feature $F_r$ at the original resolution $(H, W)$, and PE encoder feature $F_p$ at a lower resolution $(H/8, W/8)$. These features are then either upsampled or downsampled to a unified set of four spatial scales: $\mathcal{S} \in \{H, H/2, H/4, H/8\}$. At each scale, we compute attention maps using a transformer-style mechanism, where $F_h$ serves as the Query, $F_r$ as the Key, and $F_p$ as the Value. This multi-scale attention computation yields a set of attention heatmaps across different spatial levels. The outputs are then fused and upsampled back to the original input resolution. Finally, we identify the region with the highest attention response as the predicted bleeding source $(x^*, y^*)$, as illustrated in the attention heatmaps in Figure~\ref{fig:detection}C. The calculation formula is as follows:
\begin{equation}
(x^*, y^*) = \arg\max_{(x, y)} \left[ \sum_{s \in \mathcal{S}} f_{up}\left( \text{Attn}\left(F_h^s,\ F_r^s,F_p^s \right) \right) \right].
\end{equation}

\subsubsection{Detection Loss}

\textbf{Temporal Localization Loss}
The temporal localization loss is designed as a unified formulation that adaptively supervises the model based on the presence or absence of bleeding onset frames within the sliding window. Specifically, for windows without bleeding frames, only the confidence prediction is penalized to encourage a low confidence score. For windows containing bleeding frames, both the confidence score and the frame index prediction are jointly optimized to accurately localize the initial bleeding frame during the bleeding event. We use Binary Cross Entropy (BCE) loss~\cite{bishop2006pattern} to calculate the confidence loss and Mean Squared Error (MSE) loss~\cite{bishop2006pattern} to calculate the loss of the proportion within the prediction window width. We use Binary Cross Entropy (BCE) loss~\cite{bishop2006pattern} to calculate the confidence loss and Mean Squared Error (MSE) loss~\cite{bishop2006pattern} to calculate the loss of the proportion within the prediction window width. The specific detection of the initial bleeding frame loss function $\mathcal{L}_{{f}}$ is as follows:
\begin{equation}
\resizebox{0.9\columnwidth}{!}{$
\mathcal{L}_{\text{f}} = 
\underbrace{
\frac{1}{N_{\text{neg}}} \sum_{i=1}^{N_{\text{neg}}} \text{BCE}(S_{\text{conf}}^i, 0)
}_{\text{Non Bleeding Windows}} \ + \ 
\underbrace{
\left[
\text{BCE}(S_{\text{conf}}, 1) + \text{MSE}(N\cdot(\theta^t, \theta^{gt}))
\right]
\vphantom{\frac{1}{N_{\text{neg}}} \sum_{i=1}^{N_{\text{neg}}}},
}_{\text{Bleeding Windows}}
$}
\end{equation}
where $S_{conf}$ is the predicted confidence score, and $N\cdot\theta^t, N\cdot\theta^{gt} $ are the predicted and ground-truth bleeding frame indices in the window, respectively.

\textbf{Spatial Localization Loss}
To predict the bleeding source location, it is insufficient only to supervise the distance between the predicted point and the ground-truth location, as this lacks spatial guidance for the attention heatmap. To address this, we incorporate both real and pseudo labels and diffuse each point into a Gaussian distribution to generate soft attention supervision maps. These maps encourage the network to focus more intensively near the annotated locations and to suppress irrelevant regions. The overall heatmap regression is supervised using MSE loss, where different weights are assigned to pseudo labels and real labels to reflect their reliability. We use Huber loss~\cite{huber1992robust} to calculate the error between the predicted value and the ground truth of the bleeding source. The complete point supervision loss $\mathcal{L}_{\text{s}}$ is defined as:
\begin{equation}
\resizebox{0.9\columnwidth}{!}{$
\mathcal{L}_{\text{s}} = 
\lambda_1 \cdot \text{MSE}\left(H_{p}, H_{gt}\right)
+ \frac{\lambda_2}{N} \sum_{n=1}^{N} 
\text{MSE}\left(H_{p}^{n}, H_{\text{pseudo}}^{n}\right)
+ \delta \cdot \text{Huber}\left({P_{pred}}, {P_{gt}}\right),
$}
\end{equation}
where $H_{p}$ is the predicted attention heatmap, $H_{gt}$ is the Gaussian-blurred ground-truth map, and $H_{\text{pseudo}}^{n}$ denotes the pseudo-label heatmaps with $N$ total pseudo points. ${P_{pred}}$ and ${P_{gt}}$ represent the predicted and true bleeding source coordinates. We set $\lambda_1 = \lambda_2 = 0.5$ and $\delta = 1$ to balance the supervision of real/pseudo heat maps and the loss of coordinate regression.

\begin{figure*}[t]
\centering
\includegraphics[width=0.9\linewidth]{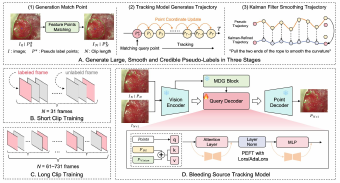}
\caption{Illustration of the BleedOrigin-Track model. 
\textbf{A} Pseudo labels are generated in three stages: (1) matching feature points across the clip, (2) generating a trajectory using a tracking model, and (3) smoothing the trajectory with a Kalman filter for temporal consistency. 
\textbf{B} and \textbf{C} show training strategies based on short clips and long clips, respectively, where only sparse labeled frames are used. 
\textbf{D} The tracking model adopts a vision encoder and query decoder with a Multi-Domain Gated (MDG) block and incorporates PEFT for lightweight fine-tuning on domain-adapted features.
}\label{fig:tracking}
\end{figure*}

\subsection{BleedOrigin-Track: Continuous Bleeding Source Tracking}

After the bleeding source is initially detected, we then track it across subsequent frames. Clinically, the bleeding source is typically located beneath the blood flow and becomes occluded when blood gushes out. However, for practical surgical guidance, this point must remain continuously visible and traceable, even under challenging conditions such as water flushing, camera jitter, or obscure bleeding view. To address this, the following two sections describe our approach for achieving robust and accurate bleeding source tracking. We detail how pseudo-labels and model-based training contribute to stable tracking performance, and how fine-tuning further enhances adaptability in dynamic surgical environments.

\subsubsection{Pseudo-label Generation and Training Phases}
We adopt Track-On~\cite{aydemir2025track}, a transformer-based point tracking framework, as the backbone for our bleeding source tracking module. Most existing transformer-based point tracking methods~\cite{karaev2024cotracker, karaev2024cotracker3,koppula2024tapvid} are trained on synthetic datasets such as MOVi-E or MOVi-F~\cite{greff2021kubric} from the Kubric datasets~\cite{greff2021kubric}. These datasets are characterized by short video lengths (typically 25 frames) and dense point annotations (e.g., 2048 sampled points per frame), which differ significantly from real surgical scenarios. In contrast, our dataset presents several challenges: it consists of real endoscopic videos with a frame rate of 30 FPS, sparse annotations (only one bleeding source per second), and video durations ranging from 5 to 10 seconds (150 to 300 frames). This difference introduces difficulties in directly applying models trained on synthetic data, motivating the need for tailored training and adaptation strategies.

As illustrated in the first stage of Figure~\ref{fig:tracking}A, we begin by generating pseudo-labels through reliable feature correspondences between two sparsely annotated frames, $I_t$ and $I_{t+30}$. Around each annotated bleeding source, we draw a circular region with radius $r$ and apply a robust feature matcher XFeat~\cite{potje2024xfeat} to extract matched keypoints within this region. We retain only those matched points that appear in both frames with matching confidence above a threshold $S$. Empirically, setting $r=50$ pixels and $S>0.7$ yields an average of approximately 36 high-confidence pseudo-keypoints $\{P_t^*\}$ per frame. The complete formula is as follows:
\begin{equation}
\resizebox{0.9\columnwidth}{!}{$
\mathcal{P}_t^{*} = \left\{
p_t^{*} \in \mathcal{N}(P_{gt}, r) \ \middle| \ 
(p_t^{*}, p_{t+30}^{*}) \in \text{Match}(I_t, I_{t+30}),\ 
S > 0.7
\right\},
$}
\end{equation}
where $\mathcal{P}_t^{*}$ denotes the set of pseudo-labeled points sampled from the neighborhood $\mathcal{N}(P_{gt}, r)$ around the ground-truth bleeding source $P_{gt}$ in the frame $I_t$. Each candidate point $p_t^{*}$ must have a valid match $p_{t+30}^{*}$ in frame $I_{t+30}$, determined by the feature matching module.

In the second stage, we track each pseudo keypoint $P_t^*$ forward for 30 frames using the pre-trained Track-On model, obtaining temporally continuous pseudo labels across intermediate frames. To refine these trajectories, the third stage applies Kalman filtering~\cite{grewal2014kalman} to reduce spatial jitter and improve robustness. Directly propagated trajectories are prone to deviation due to visual disturbances, such as motion blur or changes in illumination. However, since we already obtained reliable matched keypoints $P_{t+30}^*$ in the first step, we use them to correct the final predicted point $P_{t+30}$ and apply Kalman smoothing to the entire trajectory. The formula is as follows:
\begin{equation}
\mathcal{T}^{*} = \text{KalmanUpdate}\left( {\mathcal{T}}, \mathbf{z}_{t+30} = P_{t+30}^{*} \right),
\end{equation}
where ${\mathcal{T}} = \{P_t, P_{t+1}, \ldots, P_{t+30}\}$ represents the initially predicted trajectory obtained via Track-On, and $\mathbf{z}_{t+30} = P_{t+30}^{*}$ is the high-confidence matched keypoint used as a measurement for the Kalman update at the final frame. By incorporating this reliable endpoint observation, the Kalman filter performs backward smoothing to adjust all intermediate predictions in ${\mathcal{T}}$, resulting in the refined trajectory $\mathcal{T}^*$. This filtering process significantly mitigates drift and noise, producing temporally coherent pseudo-labels suitable for robust point tracking supervision.

To align with the training paradigm of traditional transformer-based point tracking methods, we propose two training strategies, along with a hybrid scheme, as illustrated in Figure~\ref{fig:tracking}B and C: (i) \textit{Short Clip Training:} Each batch consists of 31 consecutive frames sampled from a video, focusing on short-term temporal consistency. (ii) \textit{Long Clip Training:} Each batch contains a complete bleeding video clip, ranging from 61 to 301 frames, allowing the model to capture long-range temporal dependencies and motion variations across the entire clip.
By combining both training strategies, the hybrid training strategy effectively leverages local and global temporal cues to enhance tracking robustness under surgical conditions such as obscure bleeding view, water flushing, and camera jitter.

\subsubsection{Tracking Model with LoRA Fine-tuning}
To improve the robustness of the model in complex surgical bleeding scenarios, we introduce a deep red-attention mechanism based on the RGB channel. Specifically, we leverage the Red-Mask Block from the detection stage to generate attention-enhanced features $F_{red}$ that emphasize bright red bleeding regions. These are concatenated with the original visual features $F_{vision}$ extracted by the Track-On~\cite{aydemir2025track} backbone and used as the Key and Value in the attention mechanism to update the point feature Query.

However, directly introducing new feature channels and performing full fine-tuning on our dataset often leads to unstable training, such as exploding gradients or NaN loss, due to significant domain shifts. This issue becomes more severe when both the input data domain is drastically changed, e.g., from synthetic datasets like MOVi-E~\cite{greff2021kubric} to real-world endoscopic bleeding source tracking scenarios, and the supervision signal is greatly reduced, such as decreasing from 2048 ground truth labels to only about 60 pseudo-labeled points. To address this, we employ Low-Rank Adaptation (LoRA)~\cite{hu2022lora} to fine-tune only the attention and MLP layers with low-rank adapters, enabling stable convergence and effective model adaptation.

\subsubsection{Bleeding Source Tracking Loss}

To supervise the keypoint tracking process, we adopt a weighted Huber loss that accounts for the different levels of reliability between real and pseudo labels. Specifically, we apply a standard Huber loss~\cite{huber1992robust} to the predicted keypoint and the ground-truth keypoint, and average the Huber loss over all pseudo-labeled keypoints. The tracking loss is defined as:
\begin{equation}
\resizebox{0.9\columnwidth}{!}{$
\mathcal{L}_{\text{track}} = 
\alpha_1 \cdot \text{Huber}\left(P_{\text{pred}}, P_{\text{gt}}\right) +
\alpha_2 \cdot \frac{1}{N} \sum_{i=1}^{N} \text{Huber}\left(P_{\text{pred}}^{i}, P_{\text{pseudo}}^{i}\right),
$}
\end{equation}
where we set $\alpha_1 = 0.6$ and $\alpha_2 = 0.4$ to control the contribution of the ground truth and pseudo labels to the overall loss function. This weighting reflects our empirical observation that emphasizing the ground truth (with a slightly higher weight) leads to more stable and reliable training performance.

\subsection{Evaluation and Deployment Modes}
We adopt two distinct strategies for evaluation and deployment. In all experiments and baseline evaluations, the detection of the initial bleeding frame and the localization of the corresponding bleeding source are assessed independently. Specifically, we do not use the model-predicted initial bleeding frame to determine the bleeding source. Instead, the ground-truth initial bleeding frame is used to evaluate the initial bleeding source. Similarly, for the tracking task, we use the ground-truth initial bleeding frame and its annotated coordinate as the starting point for tracking evaluation. This ensures a consistent and fair comparison across different methods.

In contrast, for real-world deployment in clinical settings, we adopt an end-to-end pipeline: the model first identifies the frame corresponding to the initial bleeding event, then localizes the bleeding source within that frame. This point is subsequently used as the initialization for continuous tracking. As illustrated in Algorithm~\ref{alg:bleedspotnet}, we clearly distinguish between the \textbf{evaluation} and the \textbf{deployment} mode to ensure both methodological rigor and practical applicability.

\begin{algorithm}[htbp]
\caption{Detection and tracking: Evaluation and Deployment Pipeline}
\label{alg:bleedspotnet}
\begin{algorithmic}[1]
\REQUIRE $\mathcal{V}=\{I_0, I_1, \dots, I_{T-1}\}$, ${t_{gt}}$, ${P_{gt}}$, $\text{mode} \in \{\text{Evaluation}, \text{Deployment}\}$ \\
\quad Initial bleeding frame detection method $f_{bf}$,\\
\quad Initial bleeding source detection method $f_{bp}$,\\
\quad bleeding source tracking method $f_{bt}$
\ENSURE Predicted the initial bleeding frame $I_{t_b}$ and bleeding source $P_{pred}$
\IF{$\text{mode} = \text{Evaluation}$}
    \FOR{$t = 0$ to $T-1$}
        \STATE Predict the initial bleeding frame: ${t}, S_{conf}^{(t)} \gets f_{bf}(\mathcal{V})$
        \IF{$S_{conf}^{(t)}>0.5$}
            \STATE $t_b \gets t$
            \STATE \textbf{break}
        \ENDIF
    \ENDFOR
    \STATE Detect bleeding source: $P_{pred} \gets f_{bp}(I_{t_{gt}})$
    \STATE Tracking initialize $P_{t-1} \gets P_{gt}$, $I_{t-1} \gets I_{t_{gt}}$
    \FOR{$t = t_{gt}$ to $T-1$}
        \STATE $P_t \gets f_{bt}(I_t, P_{t-1})$
    \ENDFOR
\RETURN $t_b, P_{pred}, \mathcal{P}=\{P_{gt}, P_{t}, \dots, P_{T-1}\}$
\ELSIF{$\text{mode} = \text{Deployment}$}
    \FOR{$t = 0$ to $T-1$}
        \STATE Predict the initial bleeding frame: ${t}, S_{conf}^{(t)} \gets f_{bf}(\mathcal{V})$
        \IF{$S_{conf}^{(t)}>0.5$}
            \STATE $t_b \gets t$
            \STATE \textbf{break}
        \ENDIF
    \ENDFOR
    \STATE Detect bleeding source: $P_{pred} \gets f_{bp}(I_{t_{b}})$
    \STATE Tracking initialize $P_{t-1} \gets P_{pred}$, , $I_{t-1} \gets I_{t_{b}}$
    \FOR{$t = t_{gt}$ to $T-1$}
        \STATE $P_t \gets f_{bt}(I_t, P_{t-1})$
    \ENDFOR
\RETURN $t_b, P_{pred}, \mathcal{P}=\{P_{pred}, P_{t}, \dots, P_{T-1}\}$
\ENDIF
\end{algorithmic}
\end{algorithm}

\section{Experiments and Results}
In this section, we present a comprehensive evaluation of the initial bleeding frame detection method, the point detection method, and the bleeding source tracking method on our proposed ESD dataset. This includes baseline comparisons as well as ablation studies. Each subsection separately discusses detection and tracking performance to ensure clarity and focus. Finally, we evaluate the effectiveness of our approach on full-length videos to demonstrate its applicability in real-world surgical scenarios.

\subsection{Quantitative Metrics}
\textbf{Detection Evaluation Metrics}
To assess the temporal accuracy of the initial bleeding frame detection, we adopt a frame-level evaluation protocol, following prior metrics used in temporal frame localization tasks within time-series domains~\cite{salles2024softed}. Specifically, a prediction is considered correct if the predicted frame lies within a tolerance window of $\pm$$k$ frames from the ground truth. We report the percentage of correct predictions across various thresholds: $\pm0$, $\pm1$, $\pm2$, $\pm4$,
 and $\pm8$ frames. The specific formula is as follows:
\begin{equation}
\text{Acc}_{\pm k} = \frac{1}{N} \sum_{i=1}^{N} \mathbbm{1}\left( \left| t_{gt} - t_{pred} \right| \leq k \right),
\label{eq:frame_tolerance_accuracy}
\end{equation}
where \(t_{\text{pred}}\) and \(t_{\text{gt}}\) denote the predicted and ground truth frame indices corresponding to the initial bleeding event, respectively.

In addition to individual frame-level accuracies under specific tolerance windows, we report an overall average error for the initial bleeding frame (ibf) detection task, denoted as ${Err}_{avg}^{ibf}$~\cite{schmidt2024surgical}. This metric reflects the model’s consistency across varying temporal tolerances and is calculated as follows:
\begin{equation}
{Err}_{avg}^{ibf} = \frac{1}{N} \sum_{i=1}^{N}  \left( t_{gt} - t_{pred} \right).
\label{eq:acc_avg}
\end{equation}

\textbf{Tracking Evaluation Metrics}
To assess the spatial accuracy of bleeding source tracking, we follow the standard metrics used in endoscopic arbitrary point tracking tasks~\cite{schmidt2024surgical, schmidt2025point}, including the proportion of points within a given error threshold $Err_{\leq d}$ and the mean tracking error $Err_{avg}$. Specifically, a prediction is considered correct if the Euclidean distance between the predicted point and the ground truth point is within a specified pixel threshold. We report the percentage of correctly predicted points under five distance thresholds: $\leq$10px, $\leq$25px, $\leq$50px, $\leq$75px, and $\leq100$ px. The corresponding accuracy under the threshold $d$ is defined as:
\begin{equation}
Err_{\leq d} = \frac{1}{N} \sum_{i=1}^{N} \mathbbm{1} \left( \left\| p_{pred}^i - p_{gt}^i \right\|_2 \leq d \right),
\label{eq:tracking_accuracy}
\end{equation}
where $p_{pred}^i$ and $p_{gt}^i$ represent the predicted and ground truth coordinates of the bleeding source in frame $i$, respectively.

In addition to threshold-based accuracies, we also report the average tracking error (in pixels), denoted as ${Err}_{avg}$. It is defined as:
\begin{equation}
Err_{avg} = \frac{1}{N} \sum_{i=1}^{N} \left\| p_{pred}^i - p_{gt}^i \right\|_2.
\label{eq:tracking_avg_error}
\end{equation}
This metric captures the overall spatial deviation between the predicted and actual bleeding source across all evaluated frames.

\begin{figure*}[h!]
\centering
\includegraphics[width=0.7\linewidth]{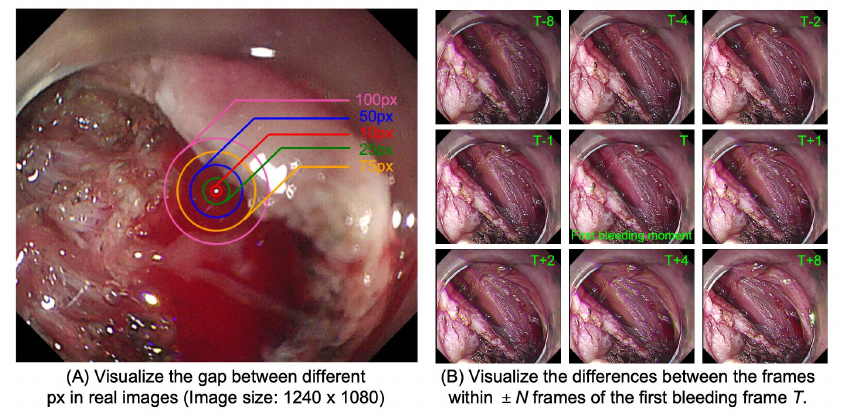}
\caption{(A) Illustration of pixel-level distances overlaid on real endoscopic frames, highlighting the spatial scale of deviations (e.g., 10px to 100px). (B) Temporal context visualization around the initial bleeding onset frame $T$, showing appearance variations within a window of $\pm N$ frames. }
\label{fig:vis_px}
\end{figure*}

\subsection{Baseline Models}
To the best of our knowledge, there are currently no available \textbf{open-source} models specifically designed for detecting the onset of bleeding events or localizing bleeding sources in ESD (see Table~\ref{tab:dataset_comp}). Therefore, we select YOLOv11~\cite{khanam2024yolov11} and YOLOv12~\cite{tian2025yolov12} as baseline architectures, each with three variants: s, n, and m. In addition, to assess the capability of cutting-edge multimodal large language models (MLLMs) in bleeding source localization, we conduct experiments using ChatGPT-4o~\cite{achiam2023gpt}, Claude 3.5~\cite{anthropic2024claude}, Gemini 1.5 Pro~\cite{team2023gemini}, and Qwen 2.5-VL~\cite{Qwen2.5-VL}.

To date, there are no existing models specifically designed for bleeding source tracking in ESD scenarios. To address this, we adopt several state-of-the-art generic point tracking models, particularly those developed under the Track-Any-Point (TAP) paradigm. These include both optical flow-based methods, including MFT~\cite{neoral2024mft} and MFTIQ~\cite{serych2025mftiq}, and Transformer-based methods, including TAPIR~\cite{doersch2023tapir}, BootsTAP~\cite{doersch2024bootstap}, CotrackerV3~\cite{karaev2024cotracker3}, and Track-On~\cite{karaev2024cotracker3}. 

We present the performance of baseline models on three subtasks: first-frame bleeding event detection, bleeding source localization, and continuous point tracking. Quantitative comparison reveals that our method achieves superior accuracy on all tasks. Figure~\ref{fig:vis_px} visualises the pixel- and frame-level discrepancies in the chosen evaluation metrics.

\subsection{Implementation Details}

For YOLOv11 and YOLOv12, we utilize the official pretrained weights corresponding to each variant and fine-tune them on our bleeding detection dataset. Each model is trained for 200 epochs using a batch size of 16. All input images are resized to \(640\times640 \), and the remaining hyperparameters are kept as their default values.

The tracking initialization uses the ground truth bleeding source coordinates rather than the detection results, ensuring fair comparison across different tracking methods and isolating tracking performance from detection. To ensure fair comparison and reproducibility, we follow the experimental protocols and implementation details established in prior work~\cite{schmidt2024surgical, schmidt2025point} on arbitrary point tracking and keypoint detection in endoscopic videos. All baseline tracking models are evaluated using their official pretrained weights and default hyperparameter configuration settings. Inference is performed directly without any additional fine-tuning or parameter modification. For our detection model, training is conducted over 500 epochs with a fixed learning rate of $1 \times 10^{-5}$. For the tracking task, we train on both short and long clips independently, each for 100 epochs with a learning rate of $5 \times 10^{-6}$. All experiments are performed on an NVIDIA A6000 GPU with 40GB of memory, which supports both model training and inference.

\begin{table*}[t]
\centering
\caption{Frame-level accuracy for detecting the initial bleeding frame under varying temporal tolerances. ${Err}_{avg}^{ibf}$ represents the average detection error across all thresholds. Temporal accuracy is reported as the percentage of predictions within $\pm k$ frames. The best performance is shown in \textbf{bold}, and the second-best is \underline{underlined}.
}
\resizebox{0.7\textwidth}{!}{
\begin{tabular}{l|ccccc|c}
\toprule
\textbf{Method} & $ \pm\leq0$ & $\pm\leq1$ & $\pm\leq2$ & $\pm\leq4$ & $\pm\leq8$ & $Err_{avg}^{ibf}$ $\downarrow$ \\
\midrule
YOLOv11-s~\cite{khanam2024yolov11} & 45.24 & \underline{54.76} & 55.52 & 69.05 & 76.19 & 5.90 \\
YOLOv11-n~\cite{khanam2024yolov11} & 44.09 & 53.26 & \underline{56.77} & 66.42 & 77.59 & 5.57 \\
YOLOv11-m~\cite{khanam2024yolov11} & 39.68 & 49.21 & 52.38 & 63.90 & 77.78 & 5.65 \\
YOLOv12-s~\cite{tian2025yolov12} & \underline{46.92} & 50.23 & 50.54 & 70.62 & \underline{86.15} & \underline{4.84} \\
YOLOv12-n~\cite{tian2025yolov12} & 40.51 & 50.67 & 58.84 & 70.83 & 83.45 & 5.11 \\
YOLOv12-m~\cite{tian2025yolov12} & 41.67 & 52.00 & 48.33 & \underline{72.67} & 83.33 & 5.21 \\
BleedOrigin-Detect-Frame (Ours) & \textbf{49.92} & \textbf{58.57} & \textbf{62.80} & \textbf{78.42} & \textbf{96.85} & \textbf{3.69} \\
\bottomrule
\end{tabular}
}
\label{tab:frame_accuracy}
\end{table*}

\subsubsection{Initial Bleeding Frame Detection}
We evaluate the performance of various models on the task of initial bleeding frame detection using frame-level tolerance-based accuracy. As shown in Table~\ref{tab:frame_accuracy}, our proposed method achieves superior performance across all temporal tolerance thresholds. Specifically, our approach yields 49.92\% accuracy at the most stringent $\pm0$-frame threshold, substantially outperforming all baseline models. The best-performing baseline, YOLOv12-s, achieves 46.92\% at this threshold, representing a 3 percentage point improvement with our method.. As the tolerance window expands, all models demonstrate improved performance; however, the relative advantage of our method becomes more evident. At the $\pm8$-frame threshold, our model achieves 96.85\% accuracy, exceeding the next-best performer (YOLOv12-s at 86.15\%) by more than 10 percentage points. These results indicate that our approach not only provides more precise bleeding onset detection but also exhibits superior robustness under relaxed temporal constraints.

Moreover, when considering the average accuracy across all tolerance levels (${Err}_{avg}^{ibf}$), our method yields the lowest error of 3.69, since this metric effectively reflects the model’s overall temporal consistency, the result indicates that our approach offers a more stable and accurate localization of the initial bleeding frame across different levels of temporal uncertainty.

\begin{table*}[t]
\centering
\caption{Performance on detecting the initial bleeding source (pixel-level accuracy). 
Accuracy is reported as the percentage of predictions within different pixel distance thresholds. $Err_{avg}$ indicates the average Euclidean distance error in pixels (lower is better). The best performance for each column is shown in \textbf{bold}, and the second-best is \underline{underlined}.}
\resizebox{0.7\textwidth}{!}{
\begin{tabular}{l|ccccc|c}
\toprule
\textbf{Method} & $\leq10$px & $\leq25$px & $\leq50$px & $\leq75$px & $\leq100$px & $Err_{avg}$ $\downarrow$ \\
\midrule
YOLOv11-s~\cite{khanam2024yolov11} & 1.16 & 11.63 & 25.58 & 37.21 & 50.00 & 146.91 \\
YOLOv11-n~\cite{khanam2024yolov11} & \underline{3.49} & 8.14  & 23.26 & 33.72 & 43.02 & 160.31 \\
YOLOv11-m~\cite{khanam2024yolov11} & 2.56 & 12.82 & 21.79 & 30.77 & 38.46 & 149.80 \\
YOLOv12-s~\cite{tian2025yolov12} & 2.33 & \underline{15.12} & \underline{39.53} & \underline{51.16} & \underline{61.63} & \underline{110.11} \\
YOLOv12-n~\cite{tian2025yolov12} & 2.33 & 10.47 & 26.74 & 38.37 & 52.33 & 132.24 \\
YOLOv12-m~\cite{tian2025yolov12} & 1.18 & 10.59 & 31.76 & 45.24  & 60.00 & 124.19 \\
\midrule
ChatGPT-4o~\cite{achiam2023gpt}     & 2.35  & 2.35  & 7.06  & 21.18 & 28.82  & 160.94 \\
Claude-3.5~\cite{anthropic2024claude}     & 0.00     & 0.00     & 9.41  & 17.65 & 32.65  & 161.54 \\
Gemini-2.5-Pro~\cite{team2023gemini} & 0.00     & 2.35  & 18.24  & 25.29 & 34.71  & 152.55 \\
Qwen2.5-VL~\cite{Qwen2.5-VL}     & 2.35  & 7.06  & 16.47 & 29.41 & 40.00  & 154.85 \\
BleedOrigin-Detect-Pixel (Ours)          & \textbf{10.25} & \textbf{15.78} & \textbf{43.50} & \textbf{58.05} & \textbf{70.24}  & \textbf{68.71}  \\
\bottomrule
\end{tabular}
}
\label{tab:pixel_accuracy}
\end{table*}

\subsubsection{Initial Bleeding Source Detection}

\textbf{Comparison with YOLO-based Detection Models} As shown in Table~\ref{tab:pixel_accuracy}, our method significantly outperforms all YOLOv11 and YOLOv12 variants across all spatial thresholds. At the strictest 10-pixel threshold, our model achieves 10.25\% accuracy, while the best-performing baseline (YOLOv 11-n) achieves only 3.49\%. The performance gap increases at larger thresholds: at 50 pixels, our method achieves 43.50\% versus 39.53\% for YOLOv12-s, and at 100 pixels, 70.24\% versus 61.63\%. Moreover, our method achieves the lowest average pixel error ($Err_{avg}$) of 68.71, demonstrating superior localization precision. In contrast, YOLOv11-n and YOLOv12-m exhibit much higher average errors of 160.31 and 124.19, respectively, indicating less stable spatial predictions. 

\begin{figure*}[!ht]
\centering
\includegraphics[width=0.9\linewidth]{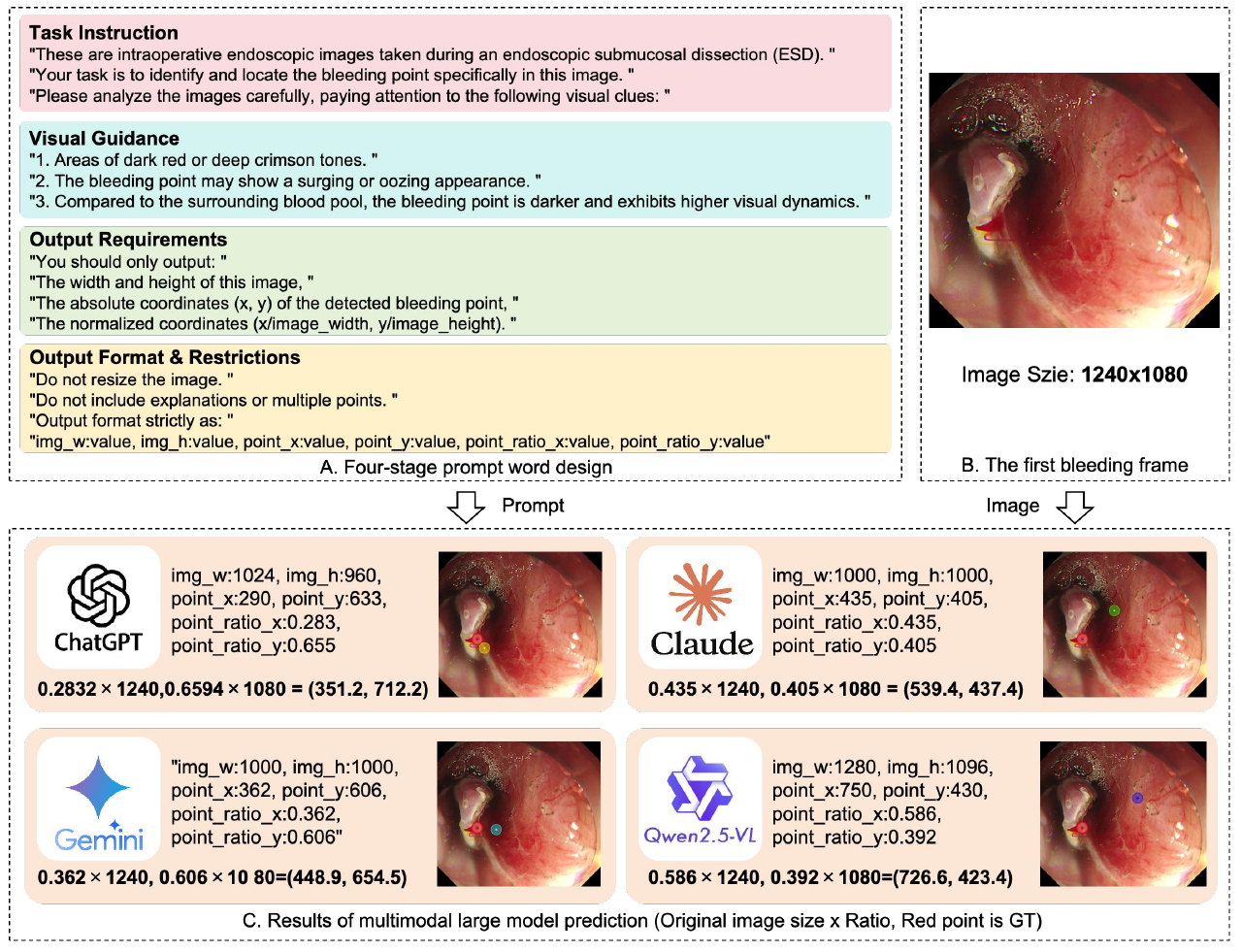}
\caption{Comparison of MLLMs' predictions for bleeding source localization in intraoperative endoscopic images.
(A) Task prompt and visual guidance provided to the models. (B) The initial bleeding frame (image size: 1240×1080) is provided to the models as the current observation. (C) Predicted bleeding source coordinates and normalized ratios from four MLMs (ChatGPT~\cite{achiam2023gpt}, Claude~\cite{anthropic2024claude}, Gemini~\cite{team2023gemini}, Qwen2.5-VL~\cite{Qwen2.5-VL}), with absolute positions calculated in the original image resolution.}
\label{fig:MLLM}
\end{figure*}

\textbf{Comparison with Multimodal Large Language Models (MLLMs)} We adopt a structured prompt design for multimodal large language models (MLLMs), as illustrated in Figure~\ref{fig:MLLM}. The prompt is composed of four components: (1) Task Instruction, (2) Visual Guidance, (3) Output Requirements, and (4) Output Format \& Restrictions. This design allows us to embed prior knowledge into the prompt, such as explicitly stating that the image is from an endoscopic submucosal dissection (ESD) procedure and describing characteristic features of the bleeding source. To ensure consistent and interpretable outputs, we define a strict format for how models should respond. During testing, we observed that different MLLMs often apply internal resizing or preprocessing to the input image. If coordinate values are returned directly in absolute pixel terms, this rescaling can introduce significant localization errors. To address this, we require models to simultaneously output both the resized image dimensions and the normalized coordinates of the predicted bleeding source. Final absolute coordinates are then computed using the original image resolution and the predicted normalized ratios.

We further evaluate four state-of-the-art MLLMs on the bleeding source localization task: ChatGPT-4o~\cite{achiam2023gpt}, Claude-3.5~\cite{anthropic2024claude}, Gemini-2.5-Pro~\cite{team2023gemini}, and Qwen2.5-VL~\cite{Qwen2.5-VL}. As shown in Table~\ref{tab:pixel_accuracy}, these models generally underperform compared to traditional detection architectures and our method. For example, ChatGPT-4o achieves only 2.35\% accuracy within 10 pixels and 7.06\% within 50 pixels, with a high average error of 160.94. Similar trends are observed for Claude-3.5 and Gemini-2.5-Pro, both of which demonstrate limited localization ability. Among the MLLMs, Qwen2.5-VL achieves the best overall performance, reaching 16.47\% accuracy at the 50-pixel threshold and 40.00\% at 100 pixels. These results suggest that current MLLMs, while impressive in general vision-language tasks, are not yet suitable for high-precision bleeding source localization without further task-specific adaptation.

\begin{table*}[ht]
\centering
\caption{Frame-level accuracy under different confidence thresholds. Percentage of predictions within $\pm k$ frames for various YOLO variants at confidence thresholds 0.25, 0.10, and 0.01. Comparative results (vs Ours) are shown in Table~\ref{tab:frame_accuracy}.}
\resizebox{\textwidth}{!}{
\begin{tabular}{l|cccccc|cccccc|cccccc}
\toprule
\multirow{2}{*}{Method} 
& \multicolumn{6}{c|}{Conf. = 0.25} 
& \multicolumn{6}{c|}{Conf. = 0.10} 
& \multicolumn{6}{c}{Conf. = 0.01} \\
\cmidrule(lr){2-7} \cmidrule(lr){8-13} \cmidrule(lr){14-19}
& $\pm\leq0$ & $\pm\leq1$ & $\pm\leq2$ & $\pm\leq4$ & $\pm\leq8$ & $\pm\leq16$ 
& $\pm\leq0$ & $\pm\leq1$ & $\pm\leq2$ & $\pm\leq4$ & $\pm\leq8$ & $\pm\leq16$
& $\pm\leq0$ & $\pm\leq1$ & $\pm\leq2$ & $\pm\leq4$ & $\pm\leq8$ & $\pm\leq16$ \\
\midrule
YOLOv11-s~\cite{khanam2024yolov11} & 0.00 & 0.00 & 0.00 & 0.00 & 0.00 & 0.00 
          & 0.00 & 0.00 & 0.00 & 0.00 & 0.00 & 0.00 
          & 45.24 & 54.76 & 59.52 & 69.05 & 76.19 & 92.86 \\
YOLOv11-n~\cite{khanam2024yolov11} & 0.00 & 0.00 & 0.00 & 0.00 & 0.00 & 0.00 
          & 0.00 & 0.00 & 0.00 & 0.00 & 0.00 & 0.00 
          & 45.24 & 54.76 & 59.52 & 69.05 & 76.19 & 92.86 \\
YOLOv11-m~\cite{khanam2024yolov11} & 0.00 & 0.00 & 0.00 & 0.00 & 0.00 & 0.00 
          & 0.00 & 0.00 & 0.00 & 0.00 & 0.00 & 0.00 
          & 39.68 & 49.21 & 52.38 & 61.90 & 77.78 & 88.89 \\
YOLOv12-s~\cite{tian2025yolov12} & 0.00 & 0.00 & 0.00 & 0.00 & 0.00 & 0.00 
          & 0.00 & 0.00 & 0.00 & 0.00 & 0.00 & 0.00 
          & 46.92 & 50.23 & 50.54 & 70.62 & 86.15 & 89.23 \\
YOLOv12-n~\cite{tian2025yolov12} & 10.34 & 13.79 & 27.59 & 41.38 & 58.62 & 75.86 
          & 31.51 & 49.32 & 56.16 & 67.12 & 82.19 & 87.67 
          & 40.51 & 50.67 & 58.84 & 70.83 & 83.45 & 100.00 \\
YOLOv12-m~\cite{tian2025yolov12} & 0.00 & 0.00 & 0.00 & 0.00 & 0.00 & 0.00 
          & 0.00 & 0.00 & 0.00 & 0.00 & 0.00 & 0.00 
          & 41.67 & 52.00 & 48.33 & 72.67 & 83.33 & 86.67 \\
\bottomrule
\end{tabular}
}
\label{tab:frame_accuracy_multi_conf}
\end{table*}

\begin{table*}[ht]
\centering
\caption{Pixel accuracy under different confidence thresholds. Percentage of predictions within $\pm k$ frames for various YOLO variants at confidence thresholds 0.25, 0.10, and 0.01.  Comparative results (vs Ours) are shown in Table~\ref{tab:pixel_accuracy}.}
\resizebox{\textwidth}{!}{
\begin{tabular}{l|cccccc|cccccc|cccccc}
\toprule
\multirow{2}{*}{Method} 
& \multicolumn{6}{c|}{Conf. = 0.25} 
& \multicolumn{6}{c|}{Conf. = 0.10} 
& \multicolumn{6}{c}{Conf. = 0.01} \\
\cmidrule(lr){2-7} \cmidrule(lr){8-13} \cmidrule(lr){14-19}
& $\pm\leq0$ & $\pm\leq1$ & $\pm\leq2$ & $\pm\leq4$ & $\pm\leq8$ & $\pm\leq16$ 
& $\pm\leq0$ & $\pm\leq1$ & $\pm\leq2$ & $\pm\leq4$ & $\pm\leq8$ & $\pm\leq16$
& $\pm\leq0$ & $\pm\leq1$ & $\pm\leq2$ & $\pm\leq4$ & $\pm\leq8$ & $\pm\leq16$ \\
\midrule
YOLOv11-s~\cite{khanam2024yolov11} & 0.00 & 0.00 & 0.00 & 0.00 & 0.00 & 0.39 
          & 0.00 & 0.00 & 0.00 & 0.00 & 0.00 & 2.14 
          & 1.16 & 11.63 & 25.58 & 37.21 & 50.00 & 146.91 \\
YOLOv11-n~\cite{khanam2024yolov11} & 0.00 & 0.00 & 0.00 & 0.00 & 0.00 & 0.12 
          & 0.00 & 0.00 & 0.00 & 0.00 & 0.00 & 2.77 
          & 3.49 & 8.14  & 23.26 & 33.72 & 43.02 & 160.31 \\
YOLOv11-m~\cite{khanam2024yolov11} & 0.00 & 0.00 & 0.00 & 0.00 & 0.00 & 0.22 
          & 0.00 & 0.00 & 0.00 & 0.00 & 0.41 & 2.73 
          & 2.56 & 12.82 & 21.79 & 30.77 & 38.46 & 149.80 \\
YOLOv12-s~\cite{tian2025yolov12} & 0.00 & 0.00 & 0.00 & 0.00 & 0.21 & 1.32 
          & 0.00 & 0.00 & 0.00 & 0.00 & 0.21 &1.48 
          & 2.33 & 15.12 & 39.53 & 51.16 & 61.63 & 110.11 \\
YOLOv12-n~\cite{tian2025yolov12} & 0.00 & 0.00 & 0.00 & 0.00 & 0.21 & 1.77 
          & 0.00 & 0.00 & 0.00 & 0.00 & 0.08 & 2.45 
          & 2.33 & 10.47 & 26.74 & 38.37 & 52.33 & 132.24 \\
YOLOv12-m~\cite{tian2025yolov12} & 0.00 & 0.00 & 0.00 & 0.00 & 0.72 & 2.89 
          & 0.00 & 0.00 & 0.00 & 0.00 & 0.31 & 3.28 
          & 1.18 & 10.59 & 31.76 & 45.24 & 60.00 & 124.19 \\
\bottomrule
\end{tabular}
}
\label{tab:px_accuracy_multi_conf}
\end{table*}

\subsubsection{The Impact of Confidence on YOLO's Bleeding Source Detection}

From Tables~\ref{tab:frame_accuracy_multi_conf} and~\ref{tab:px_accuracy_multi_conf}, it is evident that all YOLO~\cite{khanam2024yolov11, tian2025yolov12} variants perform poorly at the default confidence threshold of 0.25, yielding near-zero accuracy in both frame-level and pixel-level evaluations. Even at a relaxed threshold of 0.10, performance remains marginal. Only when the threshold is reduced to an unusually low value of 0.01 do the models begin to show localization results of the initial bleeding frame.

This phenomenon reveals a fundamental mismatch between the confidence scores output by YOLO and the requirements of bleeding onset detection. In theory, higher confidence scores should correlate with more temporally accurate and semantically reliable predictions. However, our empirical results demonstrate the opposite: predictions with high confidence are often temporally misaligned with the actual bleeding onset, leading to a complete failure under standard thresholds (e.g., 0.25). Only when the threshold is reduced to an extremely low level (e.g., 0.01) do the models begin to exhibit usable frame-level or pixel-level accuracy. This indicates that genuinely informative detections are being assigned low confidence by the model, an unintuitive and undesirable behavior. Such miscalibration undermines the practical utility of the model, as operating at low thresholds increases the risk of false positives and severely weakens the model’s discriminative power.

\begin{table*}[t]
\centering
\caption{Pixel-level accuracy of BleedOrigin-Track model. Tracking performance is reported as the percentage of predicted points within various distance thresholds from the ground truth. $Err_{avg}$ denotes the average pixel error across all frames (lower is better). }
\resizebox{0.7\textwidth}{!}{
\begin{tabular}{l|ccccc|c}
\toprule
\textbf{Method} & $\leq$10px & $\leq$25px & $\leq$50px & $\leq$75px & $\leq100$ px & $Err_{avg}$ $\downarrow$ \\
\midrule
TAPIR~\cite{doersch2023tapir}       & 7.43  & 30.93 & 52.43 & 65.90 & 76.80 & 79.77 \\
BootsTAP~\cite{doersch2024bootstap}    & 12.85 & 37.07 & 56.49 & 69.95 & 78.81 & 72.43 \\
CotrackerV3~\cite{karaev2024cotracker3}            & 13.51 & 44.86 & \underline{71.89} & \underline{85.41} & \underline{92.43} & 50.54 \\
MFT~\cite{neoral2024mft}                    & 9.41  & 36.56 & 61.83 & 74.19 & 81.72 & 68.15 \\
MFTIQ~\cite{serych2025mftiq}                  & 15.43 & 40.75  & 64.06  & 75.24 & 83.77 & 63.60 \\
Track-On~\cite{aydemir2025track}       & \underline{34.04} & \underline{53.93} & 69.94 & 78.73 & 85.60 & \underline{47.82} \\
BleedOrigin-Track (Ours)                    & \textbf{41.17} & \textbf{61.28} & \textbf{82.39} & \textbf{92.75} & \textbf{96.11} & \textbf{37.19} \\
\bottomrule
\end{tabular}
}
\label{tab:tracking_accuracy}
\end{table*}

\begin{figure*}[ht]
\centering
\includegraphics[width=1.0\linewidth]{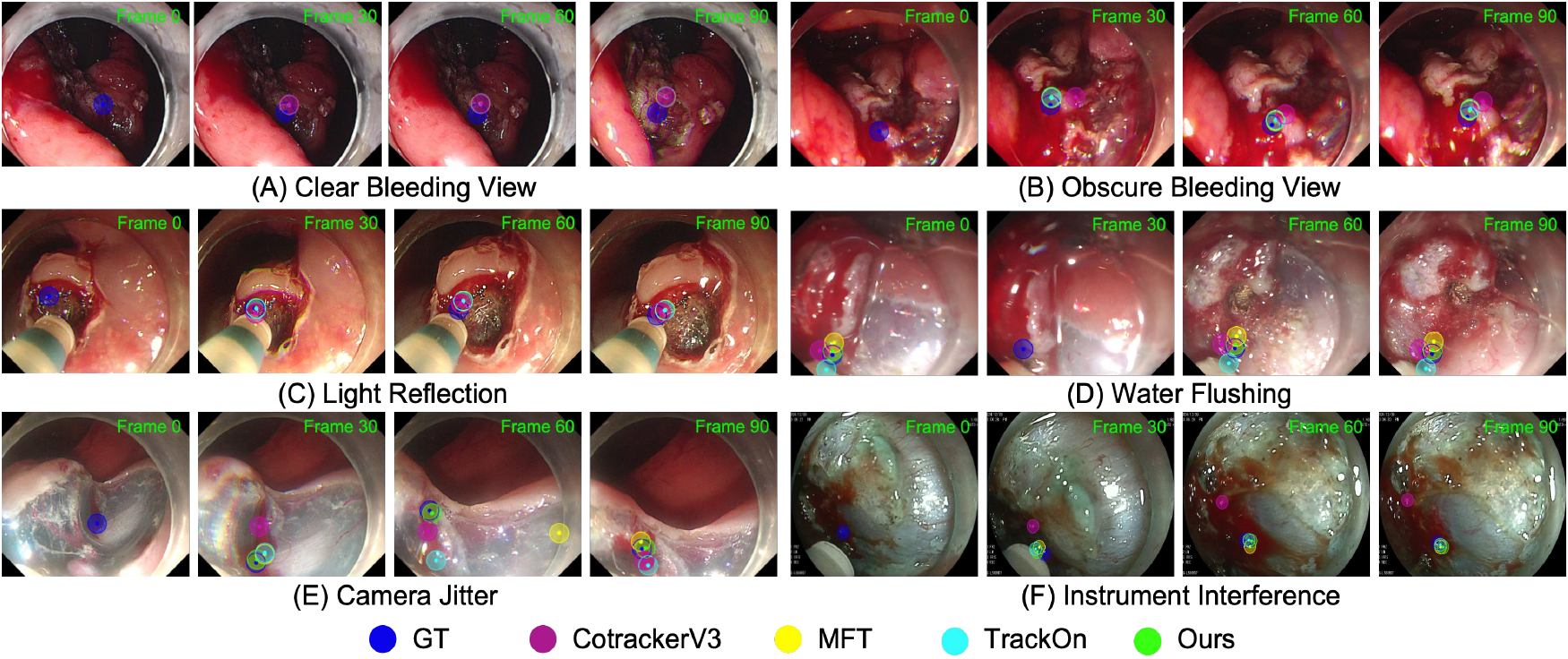}
\caption{
Qualitative comparison of bleeding source tracking across six different scenarios: (A) clear bleeding view, and five challenging conditions: (B) obscure bleeding view; (C) light reflection; (D) water flushing; (E) camera jitter; (F) instrument interference. Ground truth and predictions from CotrackerV3, MFT, TrackOn, and our method are compared, demonstrating superior performance under challenging ESD surgical conditions. The visualization results can be viewed on our ~\href{https://szupc.github.io/ESD_BleedOrigin/\#challenge_vis}{homepage}.
}
\label{fig:tracking_vis}
\end{figure*}

\subsubsection{Bleeding Source Tracking}

Our proposed method outperforms all baselines by a clear margin at every threshold. Table~\ref{tab:tracking_accuracy} presents the performance of state-of-the-art point tracking models on the bleeding source tracking task, measured in terms of pixel-level accuracy across varying distance thresholds and average localization error ($Err_{avg}$). 

Specifically, at the strictest 10-pixel threshold, our method achieves 41.17\% accuracy, significantly surpassing the second-best performer, Track-On (34.04\%). The gap continues to widen at more relaxed thresholds: at 50 pixels, our model reaches 82.39\%, compared to 71.89\% by CotrackerV3. At 100 pixels, our model attains 96.11\%, highlighting its robustness in spatial continuity. In terms of overall localization precision, our method yields the lowest average tracking error of 37.19 pixels. The closest competitor, Track-On, reports a higher $Err_{avg}$ of 47.82. These results demonstrate that our method not only provides accurate bleeding source predictions at the frame level but also exhibits superior consistency in temporal point propagation across clips.

To further assess the robustness of bleeding source tracking models, we present qualitative comparisons under five representative clinical scenarios, as shown in Figure~\ref{fig:tracking_vis}. These include: (A) clear bleeding view, (B) obscure bleeding view, (C) camera jitter, (D) light reflection, (E) water flushing, and (F) instrument interference. Each row shows four frames sampled from a clip, with predicted bleeding sources overlaid from four tracking methods: CotrackerV3, MFT, Track-On, and our method. Ground truth (GT) annotations are shown in purple.

Under ideal conditions (A), all methods can maintain reasonable tracking. However, in more challenging scenes such as (B) obscure bleeding view and (C) camera jitter, baseline methods often drift toward blood pools or edges. In contrast, our method maintains accurate localization aligned with the GT. In particular, CotrackerV3 and MFT exhibit noticeable jitter or lag when subjected to fast motion or sudden appearance changes. Our method shows strong resilience under adverse visibility conditions. In cases of (D) light reflection, (E) water flushing, and (F) instrument interference, other methods frequently lose track or mislocalize, while our model remains consistently close to the GT across all frames. The qualitative results align with our quantitative findings, highlighting the stability and precision of our method in real-world ESD tracking scenarios.

\begin{table*}[ht]
\centering
\caption{Ablation study of BleedOrigin-Detect for initial bleeding frame and point localization. \textbf{Top:} Frame-level accuracy under different temporal tolerances ($\pm k$ frames). $Acc_{avg}$ denotes the average accuracy across all thresholds. \textbf{Bottom:} Pixel-level accuracy for bleeding source localization, evaluated by the percentage of predictions within various spatial distance thresholds. $Err_{avg}$ represents the average Euclidean distance error in pixels.}
\label{tab:bleeding_frame_point}
\resizebox{0.7\textwidth}{!}{
\begin{tabular}{l|ccccc|c}
\toprule
\multicolumn{7}{c}{\textbf{Detecting the Initial Bleeding Frame (Frame-level Accuracy)}} \\
\midrule
\textbf{Memory Block} & $\pm\leq 0$ & $\pm\leq 1$ & $\pm\leq 2$ & $\pm\leq 4$ & $\pm\leq 8$ & $Err_{avg}^{ibf}$ $\downarrow$ \\
\midrule
\ding{55}   & 44.70  & 51.41  & 59.43  & 62.50  & 80.78 & 4.77 \\
$\checkmark$ &  \textbf{49.92} & \textbf{58.57} & \textbf{62.80} & \textbf{78.42} & \textbf{96.85} & \textbf{3.69} \\
\midrule
\multicolumn{7}{c}{\textbf{Detecting the Initial bleeding source (Pixel-level Accuracy)}} \\
\midrule
\textbf{Vision Encoder Block} & $\leq 10\,\mathrm{px}$ & $\leq 25\,\mathrm{px}$ & $\leq 50\,\mathrm{px}$ & $\leq 75\,\mathrm{px}$ & $\leq 100\,\mathrm{px}$ & $Err_{\text{avg}} \downarrow$ \\
\midrule
DINOV2~\cite{oquab2023dinov2}         & \underline{4.50}   & \underline{11.80}  & \underline{37.85} & \underline{39.84} & \underline{55.17} & \underline{72.90} \\
PE-Core-448~\cite{bolya2025PerceptionEncoder}    & 3.23  & 9.12  & 25.08 & 35.24 & 49.91 & 86.04 \\
PE-Lang-448~\cite{bolya2025PerceptionEncoder}    & 2.95  & 7.42  & 24.59 & 33.77 & 45.36 & 90.73 \\
PE-Spatial-448~\cite{bolya2025PerceptionEncoder} & \textbf{10.25} & \textbf{15.78} & \textbf{43.50}  & \textbf{58.05} & \textbf{70.24} & \textbf{68.71} \\
\bottomrule
\end{tabular}
}
\end{table*}

\begin{figure}[t]
\centering
\includegraphics[width=\linewidth]{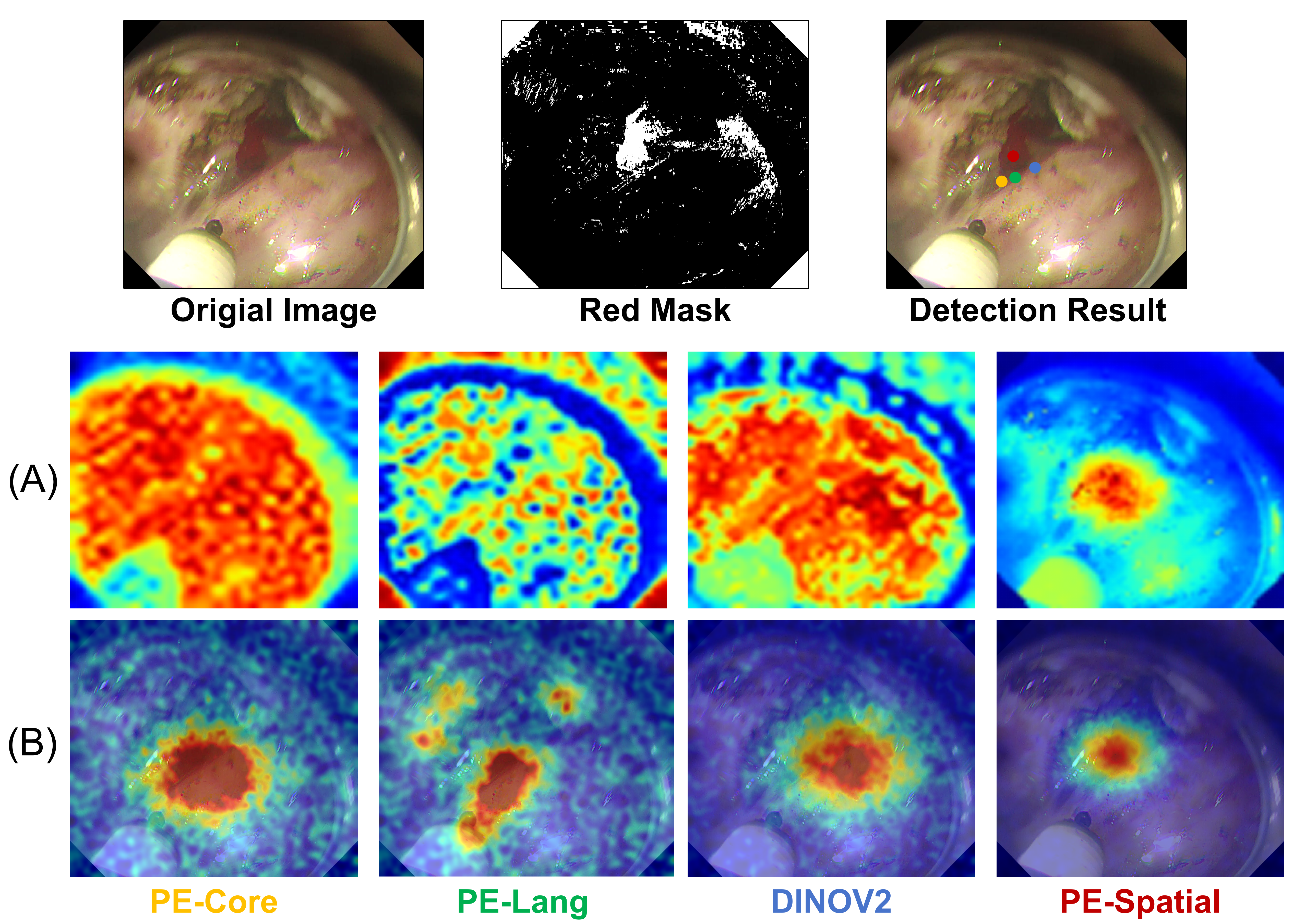}
\caption{Visual comparison of attention maps from different vision encoders for bleeding source localization. Top: original input image, red region mask, and detection result. Bottom: (A) is the feature extraction heatmaps, while (B) is the attention feature maps after final feature fusion from PE-Core, PE-Lang, DINOV2, and PE-Spatial.}
\label{fig:detection_heatmap}
\end{figure}

\subsection{Ablation Study}

To better understand the contributions of individual components in our proposed framework, we conduct comprehensive ablation studies focusing on both architectural design and training strategies. Specifically, we analyze the role of temporal memory and vision encoder backbones in the detection stage, and investigate the effects of pseudo-label supervision, temporal clip selection, chromatic attention guidance, and fine-tuning paradigms in the tracking stage. The following sections present detailed quantitative and qualitative comparisons to reveal the effectiveness of each design choice.

\subsubsection{BleedOrigin-Detect Components}

Table~\ref{tab:bleeding_frame_point} summarizes the ablation study results for two key architectural components: the memory block (used for temporal context modeling) and the vision encoder backbone (used for point localization).

\textbf{Effect of the Memory Block on Frame-level Accuracy.} Removing the memory block significantly degrades performance in the initial bleeding frame detection. Without memory, accuracy at the strictest $\pm0$ threshold drops from 49.92\% to 44.7\%, and the overall average error ($Err_{avg}^{ibf}$) deteriorates from 4.77 to 3.69. This suggests that temporal information plays a critical role in precisely identifying the onset of bleeding, and that the memory block effectively captures inter-frame dependencies to enhance temporal localization.

\textbf{Effect of Vision Encoder on Pixel-level Accuracy.}
We perform an ablation study on 4 different visual encoders to investigate their impact on bleeding source localization accuracy. The results are shown in the bottom part of Table~\ref{tab:bleeding_frame_point}. Among the compared encoders, PE-Spatial-448 significantly outperforms all others across all distance thresholds. Specifically, it achieves 10.25\% accuracy within 10 pixels, nearly doubling the performance of the next best encoder (DINOV2 at 4.5\%). At broader thresholds such as 50 and 100 pixels, PE-Spatial-448 reaches 43.50\% and 70.24\%, respectively, indicating both precise and robust localization. In terms of overall localization error, PE-Spatial-448 also yields the lowest average error ($Err_{\text{avg}}$) at 68.71 pixels, showing a clear advantage over other variants such as DINOV2 (72.90), PE-Core-448 (86.04), and PE-Lang-448 (90.73). While DINOV2 provides reasonable mid-range accuracy (e.g., 37.85\% at 50px), its performance drops significantly at higher thresholds and underperforms in fine-grained localization.

\textbf{Visual Comparison of Vision Encoders.}
Also, the Figure~\ref{fig:detection_heatmap} presents qualitative visualizations comparing attention responses from different vision encoders on the bleeding source localization task. The top row shows the input image, a red-region mask generated from color thresholding, and the final detection result overlaid with predicted coordinates. The lower rows display encoder-specific attention heatmaps, illustrating where each model focuses during prediction.

Compared to PE-Core, PE-Lang, and DINOV2, the PE-Spatial encoder produces the sharply concentrated and spatially consistent attention around the bleeding source. Its activation map aligns closely with the ground truth bleeding region, demonstrating a clear ability to isolate the bleeding source from surrounding tissues, blood pools, and visual noise. In contrast, PE-Core and PE-Lang exhibit broad and scattered attention across unrelated areas of mucosal tissue and shadow artifacts, failing to localize the small, dynamic bleeding region precisely. DINOV2 shows a moderate spatial response, but its attention is often diluted across broader reddish regions, leading to inaccurate coordinate estimation.

This difference is especially critical in our task setting, which requires predicting a precise point coordinate rather than coarse region segmentation. Unlike tasks that tolerate broader heatmaps (e.g., saliency detection), coordinate prediction demands tight, high-confidence activation centered on the true bleeding source. The PE-Spatial encoder’s ability to enhance spatial discrimination directly contributes to reduced localization error, as confirmed by its lowest $Err_{\text{avg}}$ in Table~\ref{tab:bleeding_frame_point}. These findings highlight the necessity of spatially aware visual representation learning for achieving fine-grained localization in surgical video understanding.

\begin{table*}[t]
\centering
\caption{Ablation study of fine-tuning strategies and design choices in BleedOrigin-Track. Pixel-level accuracy is reported under multiple distance thresholds, along with average localization error ($Err_{avg}$). We evaluate the impact of pseudo-label supervision, short/long temporal clips, the MDG (Motion-Dynamics-Guided) module, and PEFT methods (LoRA/AdaLoRA).}
\renewcommand{\arraystretch}{1.3}
\resizebox{0.9\textwidth}{!}{
\begin{tabular}{cccccl|ccccc|c}
\toprule
\textbf{Pseudo Labels} & \textbf{Short Clip} & \textbf{Long Clip} & \textbf{MDG Module} & \textbf{PEFT Method} & &
$\leq$10px & $\leq$25px & $\leq$50px & $\leq$75px & $\leq100$ px & $Err_{avg}$ $\downarrow$ \\
\midrule
\multicolumn{12}{c}{\textbf{Full Fine-Tuning}} \\
\midrule
\ding{55} & \checkmark & \ding{55} & \ding{55} & ---     & & 17.32 & 32.41 & 38.45 & 42.81 & 60.32 & 82.10 \\
\checkmark & \checkmark & \ding{55} & \ding{55} & ---     & & 37.63 & 54.06 & 74.80 & 86.43 & 87.04 & 41.13 \\
\checkmark & \ding{55} & \checkmark & \ding{55} & ---     & & {37.82} & {57.14} & {77.37} & {88.47} & {93.73} & 43.20 \\
\checkmark & \checkmark & \checkmark & \ding{55} & ---     & & {38.36} & {58.27} & {78.02} & {90.90} & {94.30} & 43.20 \\
\checkmark & \checkmark & \checkmark & \checkmark & ---     & & 39.02 & 59.50 & 79.42 & 86.13 & 89.75 & {45.19} \\
\midrule
\multicolumn{12}{c}{\textbf{Parameter Efficient Fine-Tuning}} \\
\midrule
\checkmark & \checkmark & \checkmark & \checkmark & LoRA~\cite{hu2022lora}    & & \underline{39.54} & \underline{60.88} & \underline{81.23} & \underline{91.80} & \underline{95.96} & \underline{39.71} \\
\checkmark & \checkmark & \checkmark & \checkmark & AdaLoRA~\cite{zhang2023adalora} & & \textbf{41.17} & \textbf{61.28} & \textbf{82.39} & \textbf{92.75} & \textbf{96.11} & \textbf{37.19} \\
\bottomrule
\end{tabular}
}
\label{tab:peft_blocks}
\end{table*}

\subsubsection{BleedOrigin-Track: Training and Fine-tuning Strategy Selection}

Table~\ref{tab:peft_blocks} presents an ablation study on several key design choices for BleedOrigin-Track, including pseudo label supervision, temporal clip structures, color-guided attention, and fine-tuning strategies.

\textbf{Effect of Pseudo Labels.} Introducing pseudo labels from weak annotations significantly improves pixel-level accuracy. Without pseudo label supervision, the model only achieves 17.32\% accuracy within 10 pixels and an average error of 82.10. Once pseudo labels are introduced, performance improves drastically, reaching over 37\% at 10 pixels and reducing $Err_{avg}$ to nearly half. This confirms that pseudo-supervised signals can effectively guide the model in early-stage localization.

\textbf{Short and Long Clip Combination.}
Training with both short and long video clips leads to further performance improvements across all pixel-level accuracy thresholds. As shown in Table~\ref{tab:peft_blocks}, using only short clips achieves 37.63\% accuracy within 10 pixels and 87.04\% within 100 pixels, with an average error of 41.13. When switching to long clips alone, we observe gains at broader thresholds, accuracy at 100 pixels increases to 93.73\%, though performance at 10 pixels remains similar (37.82\%). The notable improvement occurs when short and long clips are combined. Under this setting, accuracy increases to 38.36\% at 10 pixels and 94.30\% at 100 pixels, showing consistent gains across the range. This result confirms that short clips help with fine-grained spatial discrimination due to dense local supervision, while long clips improve robustness in temporally extended motion. Their combination allows the model to better generalize across both rapid point shifts and longer-term displacement trajectories, which are commonly observed in real ESD surgery videos.

\textbf{Effect of MDG Block.} We embed the MDG module trained in detection into the tracking model, which enables the model to attend to chromatic features such as surging redness and local bleeding contrast, improving fine-grained localization. A modest boost is observed at tight thresholds ($\leq$10px: +0.66\%), and mid- to large-scale localization consistency also benefits. However, we hypothesized that the limited improvement is due to drastic parameter shifts during full fine-tuning. To address this, we adopted LoRA~\cite{hu2022lora} as an alternative, enabling parameter-efficient fine-tuning with more stable convergence.

\textbf{Impact of Parameter-efficient Fine-tuning.}
We compare full model fine-tuning with parameter-efficient fine-tuning (PEFT) using LoRA~\cite{hu2022lora} and AdaLoRA~\cite{zhang2023adalora}. Both PEFT methods outperform full fine-tuning variants, with AdaLoRA achieving the best results across all thresholds, including 41.17\% at 10 pixels and a minimum $Err_{avg}$ of 37.19. This performance gain can be attributed to two main factors. First, the tracking model is pre-trained on datasets with substantially different distributions (e.g., MOVI-E~\cite{greff2021kubric}), and directly fine-tuning all weights on the ESD bleeding source dataset leads to unstable convergence. Second, the introduction of new trainable parameters via LoRA/AdaLoRA, focused specifically on query-key attention adaptation, allows the model to efficiently recalibrate its spatial attention without overfitting. This leads to more stable optimization and better generalization across varying surgical contexts.

\section{Deployment Experiment}

During deployment, the model processes the full-length surgical videos of two patients. To obtain quantitative evaluation results, we invited an experienced clinician to annotate three longer clips from the two patients' surgical videos: two clips approximately 8 seconds long and one clip around 14 seconds in duration. For each clip, the bleeding source is manually labeled on every frame where bleeding occurred. These clips depict the full transition from a clean field to the onset of bleeding, capturing realistic and continuous intraoperative scenarios. We apply our complete pipeline, including initial bleeding frame detection, bleeding source coordinate detection, and bleeding source tracking, to these clips and visualize the predictions frame-by-frame to assess the robustness and temporal consistency of the model qualitatively.

\begin{table*}[t]
\centering
\caption{Comparative analysis of inference performance across the three stages of our initial bleeding frame detection, point coordinate detection, and bleeding source tracking on 512 $\times$ 384 resolution video frames.}
\label{tab:model_comparison}
\resizebox{0.8\textwidth}{!}{
\begin{tabular}{lcccccc}
\toprule
\textbf{Model} & \textbf{FPS} $\uparrow$ & \textbf{Latency (s)} $\downarrow$ & \textbf{FLOPs (G)} $\downarrow$ & \textbf{Params (M)} $\downarrow$ & \textbf{Memory (MB)} $\downarrow$ \\
\midrule
Initial Bleeding Frame Detection   & 9.73  & 0.102  & 27.45  & 107.44  & 5890  \\
Initial Bleeding Source Detection   & 6.51  & 0.153  & 43.17  & 129.03  & 3260  \\
Bleeding Source Tracking          & 11.82 & 0.084  & 90.13  & 402.82  & 1874  \\
\bottomrule
\end{tabular}
}
\label{tab:inference_speed}
\end{table*}

\subsection{Parameters and Efficiency Evaluation of Model Deployment}
We evaluate the proposed method in terms of model complexity and computational efficiency on two key tasks: detecting the Initial bleeding frame and its corresponding location, and tracking the bleeding source across subsequent frames. Table~\ref{tab:inference_speed} presents the computational efficiency of our proposed three-stage pipeline, including frame-level bleeding detection, point-wise localization, and query-guided tracking. All models are evaluated using 512 $\times$ 384 resolution video frames. Notably, the tracking module achieves 11.82 FPS with a latency of only 0.084 seconds per frame, despite its relatively high FLOPs (90.13G) and parameter count (402.82M). The frame and point detection stages also maintain acceptable runtimes of 9.73 FPS and 6.51 FPS, respectively. These results indicate that the entire pipeline is capable of near real-time processing on standard GPU hardware. The efficiency also demonstrates the practical feasibility of deploying the method in real-world diagnostic workflows.

\begin{figure*}[t]
\centering
\includegraphics[width=0.9\linewidth]{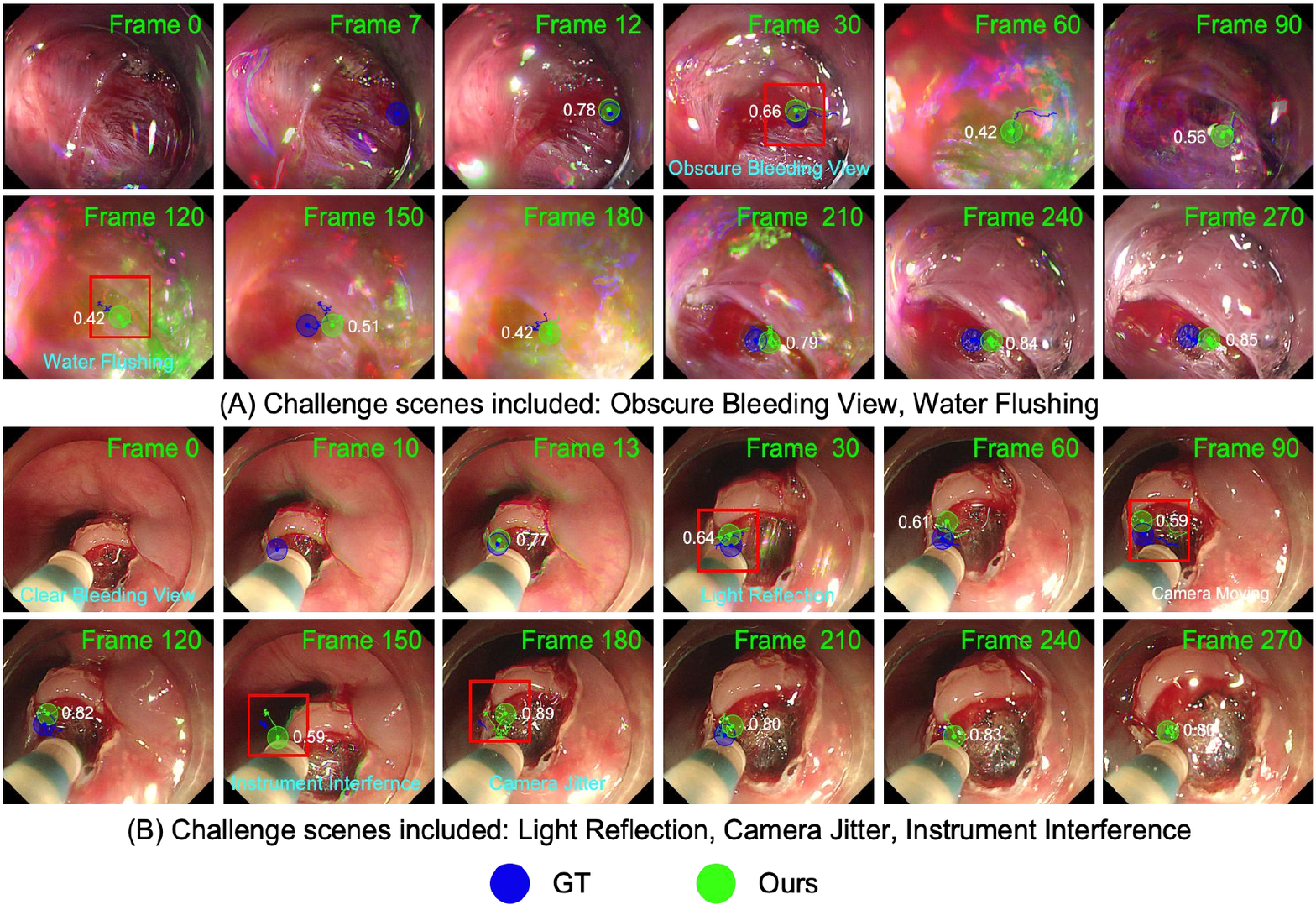}
\caption{
Visualization of our detection and tracking results on full-length ESD videos. Frame indices are shown in the top-right corners. In (A) and (B), the second image marks the ground truth of the initial bleeding frame and point, while the third shows our predicted result. The remaining frames illustrate tracking under five challenging scenarios. The white text next to each point indicates the confidence of the tracking model in predicting the point's position. The visualization results can be viewed on our ~\href{https://szupc.github.io/ESD_BleedOrigin/\#long_video}{homepage}.
}
\label{fig:long_video}
\end{figure*}

\subsection{Full-clip Tracking under Challenging Conditions}

Figure~\ref{fig:long_video} illustrates the tracking results of our method on two long-form surgical videos, capturing full bleeding clips from clean view to active hemorrhage. In each frame, the ground truth bleeding source is annotated in blue, while our predicted position is shown in green. The model is evaluated under a variety of clinically challenging conditions, including \textit{obscure bleeding view}, \textit{water flushing}, \textit{light reflection}, \textit{instrument interference}, and \textit{camera jitter}.

In clip (A), our model accurately identifies the bleeding source as it emerges (frame 12), and maintains stable tracking even during heavy bleeding (frame 30) and partial occlusion (frame 60). Although minor deviations occur under water flushing (frames 150–180), the predicted trajectory remains largely aligned with the ground truth, showing resilience to visual disturbance. Frame 254 demonstrates successful recovery once the flushing ends. In clip (B), tracking begins under ideal conditions and continues accurately even during brief occlusion caused by the instrument tool. Under light reflection (frame 60) and camera jitter (frame 210), the model exhibits slight spatial drift, as the prediction moves off-center from the ground truth. This indicates sensitivity to rapid motion and lighting artifacts, which remain challenges for pixel-level tracking. 

Similarly, our bleeding source tracking algorithm provides a corresponding confidence score for each point, which is denoted by white text in the Figure~\ref{fig:long_video}. Clinicians can adjust their trust in the model's predictions based on the magnitude of these confidence values. We observed that confidence scores decrease moderately (0.59–0.42) during instrument occlusion or water flushing, whereas they remain substantially higher (0.85–0.77) in clear visual fields.

\begin{figure*}[ht]
\centering
\includegraphics[width=0.8\linewidth]{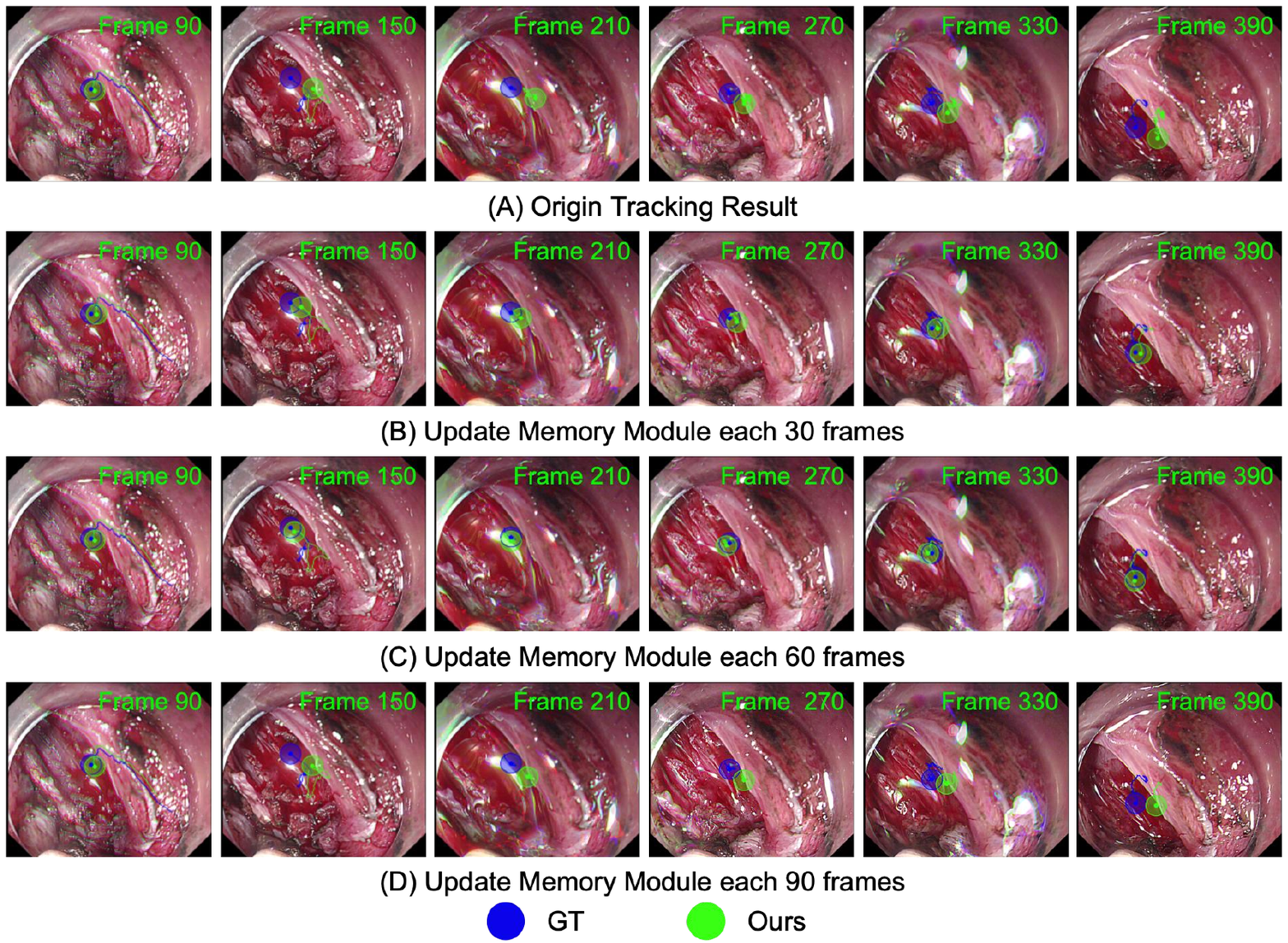}
\caption{Qualitative comparison of bleeding source tracking with different memory refresh intervals. Green denotes the predicted bleeding source, blue indicates the ground truth. Frequent memory updates improve spatial stability and reduce drift over long sequences. The visualization results can be viewed on our ~\href{https://szupc.github.io/ESD_BleedOrigin/\#update_memory}{homepage}.
}
\label{fig:memroy_restart}
\end{figure*}

\begin{figure*}[t!]
\centering
\includegraphics[width=0.8\linewidth]{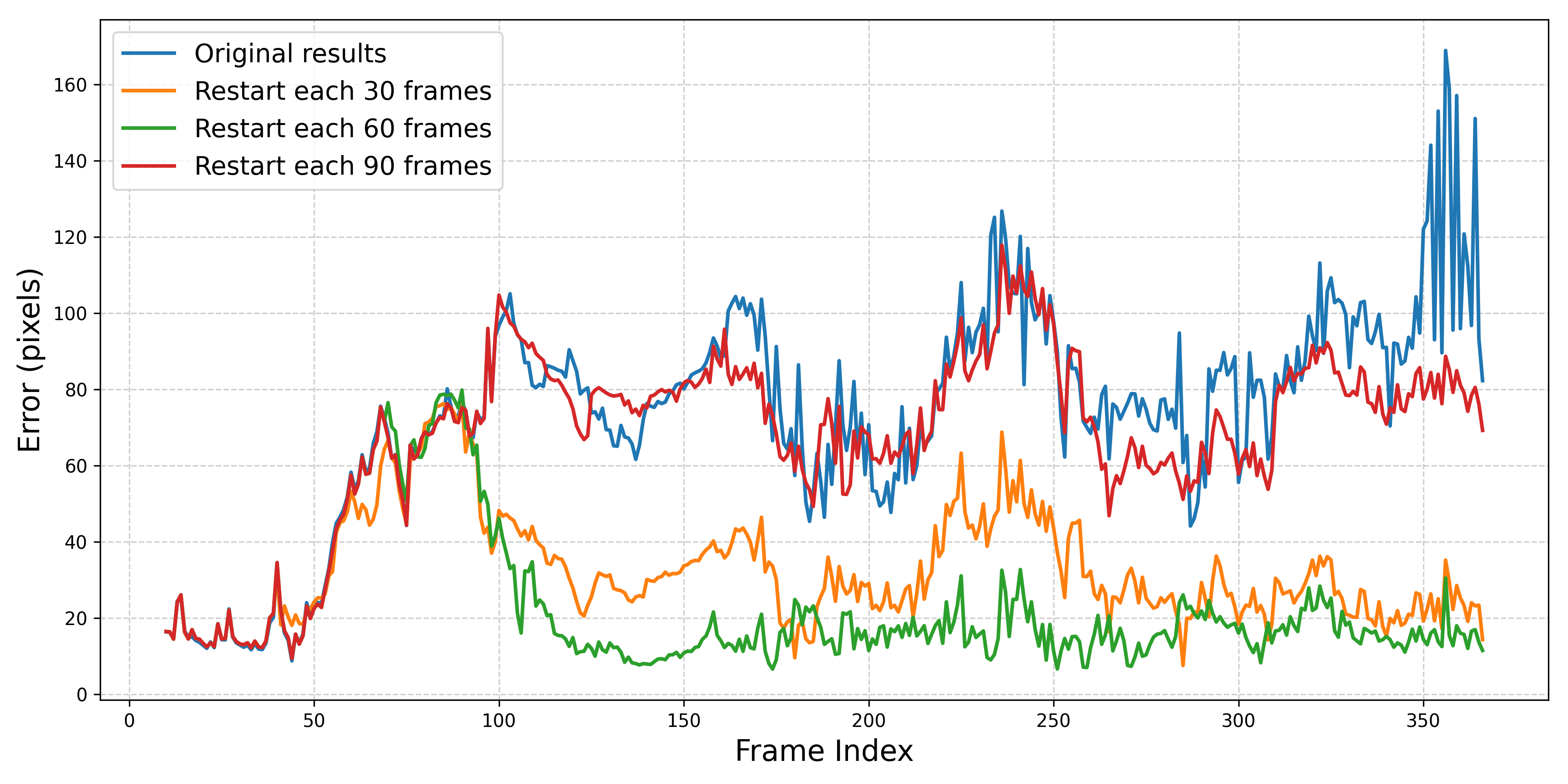}
\caption{
Frame-wise pixel error of bleeding source tracking under different memory refresh intervals. Refreshing memory every 60 frames yields the lowest and stable error, while the original strategy accumulates drift over time, and the longer the time, the greater the error accumulation and the more severe the fluctuation.}
\label{fig:restart_error}
\end{figure*}

\subsection{Effect of Memory Refresh Strategy in Long-clip Tracking}

To evaluate the impact of memory update frequency on long-term tracking performance, we experiment with periodically refreshing the memory module at fixed frame intervals. Specifically, the memory is cleared and the model is re-initialized at every $N$ frames ($N \in \{30, 60, 90\}$), using the predicted point at that frame as the new reference.

Figure~\ref{fig:memroy_restart} shows the qualitative tracking results under different refresh intervals. The original setting (A) accumulates noticeable drift and localization error over time, especially beyond 200 frames. In contrast, refreshing memory every 30 (B), 60 (C), or 90 (D) frames consistently maintains alignment with the ground truth (blue), with the 60-frame setting achieving the balance between stability and continuity. This strategy reduces error buildup caused by attention saturation or outdated contextual memory.

The quantitative pixel error plot in Figure~\ref{fig:restart_error} further validates this finding. The original Track-On model (blue curve) suffers from escalating errors as the clip progresses, exceeding 150 pixels near the end. Memory reset every 30 frames (orange) improves early tracking but introduces small fluctuations, while the 60-frame refresh (green) achieves the lowest error across the entire clip. Refreshing every 90 frames (red) helps reduce late-stage drift but lags slightly in mid-range stability.

This behavior can be attributed to the Track-On model's training setup, which is limited to short video clips (typically 25 frames), and thus lacks temporal robustness for inference on long clips exceeding 5 seconds. Transformer-based tracking architectures are generally not optimized for unbounded temporal reasoning. Our refresh strategy effectively mitigates this limitation by re-grounding the model at regular intervals, preventing cumulative drift and memory saturation. While this strategy does not constitute a fundamental architectural innovation, it serves as a highly effective inference-time enhancement for improving accuracy in long video tracking. Exploring when and how to update memory modules, especially those with dynamic or content-aware capabilities, could be a promising direction for future research.

\begin{figure*}[t!]
\centering
\includegraphics[width=0.8\linewidth]{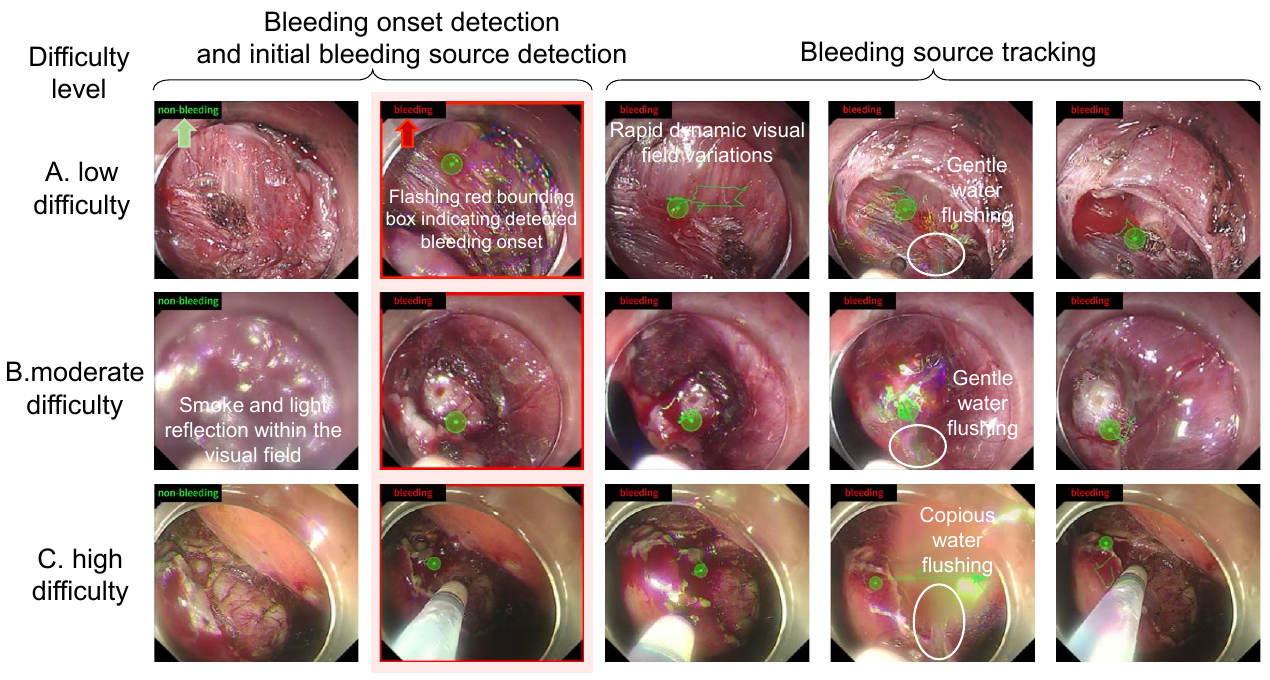}
\caption{The scene complexity progresses through three distinct tiers: low difficulty, moderate difficulty, and high difficulty. \textbf{A} Even with subtle bleeding, our model achieves timely detection via flashing red alerts and maintains accurate tracking despite rapid visual field changes; \textbf{B} Despite visibility degradation from smoke, light reflection, and gentle flushing, the model maintains both timely bleeding onset detection and stable bleeding source tracking; \textbf{C} Tracking points drift during copious flushing, but our memory refresh strategy enables bleeding source reidentification when visual features reappear post-flushing. The visualization results can be viewed on our ~\href{https://szupc.github.io/ESD_BleedOrigin/\#deployment_results}{homepage}.}
\label{fig:deployment_results}
\end{figure*}

\subsection{Clinical Feedback and User Experience Evaluation}
For deployment testing, we evaluated our BleedOrigin-Net's performance on three video clips extracted from surgical videos of two additional patients (external validation data not included in our dataset). These three complete surgical videos capture the transition from a clean field of view to active bleeding. The videos are stratified by escalating procedural difficulty: (i) Low difficulty: gentle water flushing, rapid dynamic visual field variations; (ii) Moderate difficulty: visibility is degraded by smoke and light reflection, with gentle water flushing present; (iii) High difficulty: copious water flushing causing field contamination; During the transition from the non-bleeding phase to the bleeding phase, the non-bleeding status is indicated by a ``\textbf{\textit{Non Bleeding}}'' label displayed in the upper-left corner. Upon entering the bleeding phase indicated by a ``\textbf{\textit{Bleeding}}'' label, the bleeding onset is detected and highlighted by \textbf{\textit{the flashing red bounding box}} to draw the clinician's attention to the bleeding source. Simultaneously, the detected bleeding source is displayed and continuously tracked throughout subsequent challenging environments.

Since ground truth annotations are unavailable for these clips, we conducted structured interviews with three experienced endoscopists and collected their clinical feedback to assess the clinical utility and user acceptance of our bleeding source detection and tracking system. All participating clinicians expressed willingness to use the system clinically, rating it as ``highly valuable'' for ESD procedures. 
The clinicians consistently reported ``satisfactory to excellent'' accuracy in detecting bleeding onset under normal visibility conditions, with Dr. Chaoyang Lyu noting the system's ability to identify subtle bleeding often missed during rapid dynamic visual field variations (see Figure~\ref{fig:deployment_results}A). Dr. Huifang Fan state that despite visibility degradation from smoke, light reflection, and gentle flushing, the model maintains both timely bleeding onset detection and stable point tracking. Dr. Zhen Li observed that while tracking points exhibit slight drift during copious water flushing, our memory refresh strategy enables the system to reidentify the bleeding source when clear visual features reappear after water flushing cessation (see Figure~\ref{fig:deployment_results}C). The clinical feedback validates the system's practical utility, particularly highlighting its potential to provide ``consistent support that could improve patient safety and procedural efficiency, especially for less experienced surgeons.'' By providing early AI-assisted alerts, surgeons can intervene promptly, potentially reducing dependence on repeated water flushing and improving overall procedural efficiency.

\section{Discussion and Conclusion}
This study addresses a critical gap in AI-assisted surgical safety by introducing the first comprehensive framework for bleeding source localization in Endoscopic Submucosal Dissection (ESD) procedures. Our contributions span dataset creation, methodological innovation, and clinical validation, establishing a new foundation for computational bleeding management in minimally invasive surgery.

\subsection{Key Contributions and Innovations}
\textbf{Dataset Contribution.} We introduce BleedOrigin-Bench, the first large-scale ESD bleeding source dataset comprising 44 procedures (106,222 frames) with 1,771 expert-annotated bleeding sources and 39,755 pseudo-labeled frames. The dataset captures 8 anatomical sites and 6 challenging clinical scenarios, filling a critical void in surgical AI research and enabling standardized evaluation for this essential safety task.

\textbf{Methodological Innovation.} Our BleedOrigin-Net framework addresses the complete workflow from bleeding onset detection to continuous spatial tracking through two key innovations: (1) The Multi-Domain Confidence-based Frame Memory (MDCFM) module maintains temporal context while filtering visual noise by adaptively weighting RGB, HSV, and optical flow features, enabling distinction between genuine bleeding onset and transient visual disturbances; (2) A novel pseudo-label generation pipeline combining feature matching, trajectory prediction, and Kalman filtering creates dense supervision from sparse annotations, coupled with parameter-efficient LoRA fine-tuning for stable model adaptation.

\textbf{Clinical Performance.} Our framework achieves state-of-the-art performance with 96.85\% frame-level accuracy ($\pm\leq8$ frames) for bleeding onset detection, 70.24\% pixel-level accuracy ($\leq100$ px) for initial bleeding source detection, and 96.11\% pixel-level accuracy ($\leq100$ px) for continuous tracking. Notably, our method substantially outperforms multimodal large language models (ChatGPT-4o, Claude-3.5, Gemini, Qwen2.5-VL), which achieve only $\leq$40.00\% accuracy at the 100-pixel threshold, highlighting the necessity of specialized architectures for high-precision surgical applications.

\subsection{Clinical Significance and Technical Impact}

The Multi-Domain Gated Attention mechanism provides crucial spatial guidance for bleeding source localization, effectively handling the rapid alternations between clear and blood-obscured views characteristic of ESD procedures. Our tracking performance demonstrates particular clinical value, with accuracy gains exceeding 20 percentage points at the 10-pixel threshold compared to state-of-the-art methods, enabling robust point localization even under challenging conditions such as instrument interference and water flushing.

The memory refresh mechanism addresses long-term tracking stability, which is essential for real-world deployment, while our dual deployment modes (autonomous and clinician-assisted) provide flexibility for varying clinical workflows. Clinical feedback from four experienced endoscopists validate the system's practical utility, with all participants expressing willingness for clinical adoption and three rating it as ``highly valuable'' for ESD procedures.

\subsection{Limitations and Future Directions}

Several limitations warrant consideration. Our dataset, while comprehensive for ESD procedures, originates from a single institution, potentially limiting generalizability across different surgical centers. The current approach assumes single bleeding sources per frame and operates on 2D images without depth information, which could enhance spatial precision. The memory refresh strategy, although effective, requires periodic re-initialization that interrupts continuous monitoring during critical moments.

Future research directions include: (1) Multi-institutional dataset expansion with denser temporal annotations and diverse endoscopic systems; (2) Integration of depth information through stereo endoscopy for improved spatial localization; (3) Development of adaptive memory management with dynamic refresh intervals based on visual content and tracking confidence; (4) Extension to multiple simultaneous bleeding sources with semantic surgical context understanding; (5) Integration with robotic systems for automated hemostatic intervention.

\subsection{Paradigm Shift and Clinical Impact}

This work enables a paradigm shift in surgical hemorrhage management from reactive treatment to proactive prevention. Unlike traditional approaches relying on surgeons' visual assessment, AI-driven real-time bleeding detection can identify subtle bleeding precursors, providing early warnings for timely intervention. This transition from ``detect-and-treat'' to ``predict-and-prevent'' has the potential to substantially reduce intraoperative blood loss, lower complication risks, and significantly enhance patient safety.

\subsection{Conclusion}

We have established the first comprehensive framework for bleeding source localization in ESD procedures, providing both foundational dataset contributions and methodological innovations for computational surgical assistance. The demonstrated improvements in detection and tracking accuracy, combined with clinical validation and practical deployment strategies, represent significant progress toward enhancing surgical safety in minimally invasive procedures. Our work establishes a robust foundation for future developments in AI-assisted bleeding management, with immediate implications for improving patient outcomes and reducing operative complications in ESD procedures.

\section*{Acknowledgements}
We would like to express our sincere gratitude to the clinical experts and doctors, Dr. Yang Yan from Yantaishan Hospital, Dr. Huifang Fan from Yuncheng Hospital of Traditional Chinese Medicine, Dr. Tingjuan Shi from Yuncheng Central Hospital affiliated to Shanxi Medical University, and Dr. Chaoyang Lyu from Qilu Hospital of Shandong University, for their invaluable assistance in annotating the endoscopic video frames. Their expertise and diligent work are crucial for creating the dataset and validating our results.

\bibliographystyle{elsarticle-harv} 
\bibliography{mybib}



\end{document}